\documentclass[review]{elsarticle}
\graphicspath{ {./figures/} }
\usepackage{hyperref}
\usepackage{float}
\usepackage{verbatim} 
\usepackage{apalike}
\usepackage{longtable}
\usepackage{booktabs}
\usepackage{array}

\usepackage{multirow}
\usepackage{graphicx}
\usepackage{amssymb}
\usepackage{lineno}
\usepackage{graphicx}
\usepackage{multirow}
\usepackage{dblfloatfix} 
\usepackage[dvipsnames]{xcolor}
\usepackage{soul}
\usepackage{color}
\usepackage{url}
\usepackage{float}
\usepackage{boldline}
\usepackage{xcolor,colortbl}
\usepackage{tikz}

\usepackage{pifont}
\usepackage{booktabs}
\usepackage{longtable}
\usepackage{booktabs}

\definecolor{Gray}{gray}{0.85}
\definecolor{LightCyan}{gray}{0.85}

\newcolumntype{g}{>{\columncolor{Gray}}c}

\usepackage[separate-uncertainty = true,multi-part-units = repeat]{siunitx}
\usepackage{array,amsfonts,amsmath}
\usepackage[font=small]{caption}
\usepackage{subcaption}

\usepackage[english]{babel}
\usepackage{lipsum}
\usepackage[pscoord]{eso-pic}
\usepackage{xspace}

\restylefloat{figure}
\restylefloat{table}

\journal{}

\biboptions{authoryear}

\begin{document}
\begin{frontmatter}
\begin{titlepage}
\begin{center}
\vspace*{1cm}

\textbf{ \large Unified Review and Benchmark of Deep Segmentation Architectures for Cardiac Ultrasound on CAMUS}

\vspace{1.5cm}

Zahid Ullah$^{a}$ (zahid1989@dongguk.edu), Muhammad Hilal$^{b}$ (hilal1991@sejong.ac.kr), Eunsoo Lee$^{a}$ (dmstn7432@naver.com), Dragan Pamucar$^{c, d, e}$ (dragan.pamucar@fon.bg.ac.rs), Jihie Kim$^{a^*}$ (jihie.kim@dgu.edu) \\

\hspace{10pt}

\begin{flushleft}
\small  
$^a$Department of Computer Science and Artificial Intelligence, Dongguk University, Seoul 04620, Republic of Korea \\
$^b$ Department of Semiconductor System Engineering, Sejong University, Seoul 05006, Republic of Korea \\
$^c$ Department of Operations Research and Statistics, Faculty of Organizational Sciences, University of Belgrade, Belgrade, Serbia \\
$^d$ Department of Industrial Engineering $\&$ Management, Yuan Ze University, Taoyuan City 320315, Taiwan \\
$^e$ Department of Applied Mathematical Science, College of Science and Technology, Korea University, Sejong 30019, Republic of Korea \\

\vspace{1cm}
\textbf{Corresponding Author:} \\
Jihie Kim \\
Department of Computer Science and Artificial Intelligence, Dongguk University, Seoul 04620, Republic of Korea \\
Tel: +82-2-2260-4973 \\
Email: jihie.kim@dgu.edu 

\end{flushleft}        
\end{center}
\end{titlepage}

\title{Unified Review and Benchmark of Deep Segmentation Architectures for Cardiac Ultrasound on CAMUS}

\author[label1]{Zahid Ullah}
\ead{zahid1989@dongguk.edu}

\author[label2]{Muhammad Hilal}
\ead{hilal1991@sejong.ac.kr}

\author[label1]{Eunsoo Lee}
\ead{dmstn7432@naver.com}

\author[label3,label4,label5]{Dragan Pamucar} 
\ead{dragan.pamucar@fon.bg.ac.rs}

\author[label1]{Jihie Kim \corref{cor1}
\corref{cor1}}
\ead{jihie.kim@dgu.edu}

\cortext[cor1]{Corresponding author.}
\address[label1]{Department of Computer Science and Artificial Intelligence, Dongguk University, Seoul 04620, Republic of Korea}
\address[label2]{Department of Semiconductor System Engineering, Sejong University, Seoul, 05006, Republic of Korea}
\address[label3]{Department of Operations Research and Statistics, Faculty of Organizational Sciences, University of Belgrade, Belgrade, Serbia}
\address[label4]{Department of Industrial Engineering $\&$ Management, Yuan Ze University, Taoyuan City 320315, Taiwan}
\address[label5]{Department of Applied Mathematical Science, College of Science and Technology, Korea University, Sejong 30019, Republic of Korea}

\begin{abstract}
Accurate cardiac chamber segmentation is essential for reliable estimation of left-ventricular (LV) volumes and ejection fraction (EF), yet reported performance varies widely across architectures and preprocessing choices. While several review papers summarize cardiac imaging and DL advances, few works connect this overview to a unified and reproducible experimental benchmark. In this study, we combine a focused review of cardiac ultrasound segmentation literature with a controlled comparison of three influential architectures, U-Net, Attention U-Net, and TransUNet, on the Cardiac Acquisitions for Multi-Structure Ultrasound Segmentation (CAMUS) echocardiography dataset. Our benchmark spans multiple preprocessing routes, including native NIfTI volumes, 16-bit PNG exports, GPT-assisted polygon-based pseudo-labels, and self-supervised pretraining (SSL) on thousands of unlabeled cine frames. Using identical training splits, losses, and evaluation criteria, a plain U-Net achieved a 94\% mean Dice when trained directly on NIfTI data (preserving native dynamic range), while the PNG-16-bit workflow reached 91\% under similar conditions. Attention U-Net provided modest improvements on small or low-contrast regions, reducing boundary leakage, whereas TransUNet demonstrated the strongest generalization on challenging frames due to its ability to model global spatial context, particularly when initialized with SSL. Pseudo-labeling expanded the training set and improved robustness after confidence filtering. Overall, our contributions are threefold: (i) a harmonized, apples-to-apples benchmark of U-Net, Attention U-Net, and TransUNet under standardized CAMUS preprocessing and evaluation; (ii) practical guidance on maintaining intensity fidelity, resolution consistency, and alignment when preparing ultrasound data; and (iii) an outlook on scalable self-supervision and emerging multimodal GPT-based annotation pipelines for rapid labeling, quality assurance, and targeted dataset curation. The findings offer both a concise review of the field and a practical recipe for building strong and reproducible baselines for cardiac ultrasound segmentation.
 Code is available at: \url{https://github.com/Zahid672/CAMUS_public}.
\end{abstract}

\begin{keyword}
Medical Image Analysis \sep Cardiac Image \sep Cardiacvascular Disease  \sep Deep Learning  \sep Self-supervised Learning.
\end{keyword}

\end{frontmatter}

\section{Introduction}
\label{intro}
Cardiovascular diseases (CVDs) \cite{white1987left,norris1992prognosis} remain the leading cause of death worldwide \cite{chen2020deep,li2020dual}, responsible for roughly 17.9 million deaths each year, according to the World Health Organization. In the United States, CVDs, including heart disease and stroke, continue to be the primary cause of mortality. Data from the Centers for Disease Control and Prevention (CDC) show that heart disease alone accounted for about 697,000 deaths in 2020. These figures highlight the enormous burden CVDs place on global health and emphasize the urgent need for improved medical research, early detection, and preventive healthcare strategies. Accurate diagnosis, effective risk assessment, and timely treatment are essential to reduce their impact and enhance patient survival \cite{sun2023social,update2017heart}.

Cardiac imaging includes a range of medical imaging techniques developed to visualize and evaluate the heart’s anatomy and function \cite{li2023multi}. These modalities provide detailed information about the heart’s chambers, vessels, and surrounding tissues, playing several vital roles in clinical practice. such as (i) Diagnosis and biomarkers where Cardiac imaging helps identify biomarkers associated with the risk, development, and treatment response of heart diseases \cite{vasan2006biomarkers}. For instance, carotid artery wall thickness can serve as a biomarker for atherosclerosis and help predict future cardiovascular events. It can also detect alterations in cardiac tissue or blood flow that indicate underlying heart conditions \cite{michelhaugh2023using}, thereby improving disease understanding and informing treatment decisions. (ii) Treatment planning and monitoring where Cardiac imaging is crucial in managing CVDs such as coronary artery disease (CAD), heart failure, and valvular disorders. Techniques like angiography visualize coronary blood flow and locate blockages, while echocardiography assesses heart function and detects structural or motion abnormalities. These approaches enable physicians to design effective treatment plans and make data-driven adjustments over time as patients respond to therapy. (iii) Clinical research where cardiac imaging is essential for assessing the safety and effectiveness of emerging medical treatments. It provides real-time visualization of the heart’s internal structures and functions, helping researchers monitor morphological and functional changes caused by new drugs or interventions. This ability to track cardiac responses is critical for validating therapeutic efficacy and safety before broader clinical application. Table \ref{tab:acronyms} list all acronyms used in this paper.

\begin{table}[!ht]
\centering
\caption{List of acronyms used in this paper.}
\label{tab:acronyms}
\begin{tabular}{ll}
\toprule
\textbf{Acronym} & \textbf{Description} \\
\midrule
AI   & Artificial Intelligence \\
CNN  & Convolutional Neural Network \\
DL   & Deep Learning \\
LV   & Left Ventricle \\
LA   & Left Atrium \\
RV   & Right Ventricle \\
EF   & Ejection Fraction \\
ED   & End Diastole \\
ES   & End Systole \\
CAMUS & Cardiac Acquisitions for Multi-Structure Ultrasound Segmentation \\
U-Net & U-shaped Convolutional Neural Network \\
AG   & Attention Gate \\
ViT  & Vision Transformer \\
TransUNet & Transformer-based U-Net Architecture \\
SAM  & Segment Anything Model \\
SSL  & Self-Supervised Learning \\
NIfTI & Neuroimaging Informatics Technology Initiative Format \\
PNG  & Portable Network Graphics \\
DSC  & Dice Similarity Coefficient \\
IoU  & Intersection over Union \\
HD   & Hausdorff Distance \\
ASD  & Average Surface Distance \\
GFLOPs & Giga Floating-Point Operations \\
GPU  & Graphics Processing Unit \\
\bottomrule
\end{tabular}
\end{table}

Cardiac image segmentation plays a vital role in clinical cardiology, enabling quantitative assessment of left ventricular (LV) volumes \cite{ribeiro2022left}, wall motion, and ejection fraction (EF) parameters that are fundamental for diagnosing and monitoring cardiac diseases. Manual delineation of cardiac structures is, however, time-consuming and prone to inter-observer variability \cite{petitjean2011review,tavakoli2013survey}, motivating the need for automated segmentation methods that are accurate, reproducible, and robust across patient populations and imaging modalities.

Over the past decade, deep learning (DL) has revolutionized medical image analysis. Convolutional neural networks (CNNs) have emerged as powerful feature extractors capable of learning hierarchical spatial representations directly from imaging data. Within this paradigm, U-Net \cite{ronneberger2015u} has become the canonical backbone for medical image segmentation tasks due to its encoder–decoder architecture and skip connections that preserve spatial resolution. U-Net has been successfully applied across diverse medical domains, including MRI, CT, ultrasound, and histopathology, providing high segmentation accuracy with relatively simple designs and limited data requirements.

Despite the strong baseline performance of U-Net, challenges remain in accurately segmenting small or low-contrast cardiac structures such as the apex and valve regions, where attention to localized contextual cues is critical. To address these limitations, Attention U-Net \cite{oktay2018attention} was proposed, integrating attention gates that dynamically highlight salient regions while suppressing irrelevant background activations. This allows the model to better focus on critical anatomical areas, improving delineation around complex cardiac boundaries and reducing false-positive predictions.

More recently, the integration of Vision Transformers (ViTs) into hybrid architectures has introduced the ability to model long-range dependencies, which are often crucial in capturing global cardiac context and inter-frame continuity. TransUNet \cite{chen2021transunet}, a hybrid CNN–Transformer model, combines convolutional feature extraction with Transformer-based global self-attention. The CNN encoder extracts fine-grained local features, while the Transformer encoder captures long-range spatial relationships across the entire image, resulting in improved contextual understanding and generalization.

Despite the success of these models in isolation, a critical gap remains such as most studies report their performance independently, under differing experimental setups, preprocessing pipelines, and dataset splits. This makes it difficult to assess the relative merits of U-Net, Attention U-Net, and TransUNet in a fair and consistent manner, particularly for echocardiography datasets where data characteristics, such as grayscale intensity distribution, speckle noise, and motion artifacts strongly influence network behavior.

To address this limitation, the present study provides a comprehensive and controlled comparison of three widely used architectures, U-Net, Attention U-Net, and TransUNet, on the Cardiac Acquisitions for Multi-Structure Ultrasound Segmentation (CAMUS) echocardiography dataset. All models are trained under harmonized conditions, including identical preprocessing, augmentation, and evaluation protocols. The experiments further examine the impact of data representation (NIfTI vs. 16-bit PNG conversion), pseudo-label augmentation, and self-supervised (SSL) encoder pretraining on segmentation performance and generalization.

Existing review papers such as \cite{chen2020deep,fu2021review} have provided valuable summaries of recent advances in artificial intelligence (AI) for cardiovascular imaging. However, these works are primarily narrative or taxonomic surveys, focusing on literature classification, dataset availability, and architectural trends across different cardiac modalities or diseases \cite{li2022medical,jone2022artificial}. Their emphasis lies in synthesizing the state of the field rather than performing unified experimental assessments or analyzing how preprocessing decisions affect segmentation outcomes. In contrast, the present study goes beyond descriptive synthesis by offering a systematic, controlled, and reproducible empirical benchmark of three widely used architectures, U-Net, Attention U-Net, and TransUNet, trained and evaluated under identical conditions on the CAMUS echocardiography dataset (see Table \ref{tab:review_comparison}). The pipeline enforces strict consistency in preprocessing (NIfTI vs. PNG-16-bit formats), data pairing, augmentation, loss functions, and evaluation metrics, ensuring that architectural differences, not data handling, drive performance variations. Additionally, the study introduces a Segment Anything Model (SAM) based pseudo-labeling framework that leverages unlabeled ultrasound frames to support semi-supervised learning. After confidence filtering, the SAM-generated masks provide measurable improvements, achieving Dice scores of 0.9988, 0.6694, and 0.6238 for classes C1–C3, respectively. This demonstrates the practical value of modern foundation models for enriching cardiac ultrasound datasets. 

Overall, the goal of this work is two-fold i.e., to provide an evidence-based review and a harmonized, apples-to-apples benchmark of state-of-the-art segmentation architectures on cardiac ultrasound data. To offer practical methodological guidance, spanning preprocessing fidelity, model design, data representation, and pretraining strategies, to inform and accelerate future research in cardiac ultrasound segmentation. This dual perspective positions the study as both a scholarly review and a rigorous experimental contribution, bridging the gap between theoretical surveys and hands-on reproducible evaluation. The main contributions of this work are summarized as follows:
\begin{itemize}
    \item We present a unified and reproducible benchmark of three representative deep segmentation architectures (U-Net, Attention U-Net, and TransUNet) for cardiac ultrasound segmentation on the CAMUS dataset, evaluated under strictly standardized preprocessing, training, and evaluation protocols.
    \item We systematically analyze the impact of data representation and preprocessing choices, including native NIfTI volumes versus 16-bit PNG workflows, highlighting how intensity fidelity and alignment significantly influence segmentation performance.
    \item We develop a general SAM-assisted pseudo-labeling framework that leverages unlabeled echocardiography frames through confidence-based filtering and noise-aware integration, enabling effective semi-supervised training without modifying or retraining the SAM model.
    \item We provide practical methodological guidance on architecture selection, preprocessing strategies, pseudo-label utilization, and self-supervised pretraining, offering a reusable framework that can be applied to similar medical image segmentation problems with limited annotations.
\end{itemize}


\begin{table*}[!t]
\centering
\caption{Comparison between existing review papers and the present study. Unlike prior descriptive surveys, this work provides an empirical benchmark, standardized preprocessing, and SAM-based semi-supervised learning for cardiac ultrasound segmentation.}
\resizebox{\textwidth}{!}{
\begin{tabular}{p{4cm} p{6cm} p{6cm}}
\hline
\textbf{Study} & \textbf{Scope and Focus} & \textbf{Difference from Present Work} \\ 
\hline
\textbf{Deep Learning for Cardiac Image Segmentation: A Review \cite{chen2020deep}} & 
Comprehensive overview of DL methods in cardiac segmentation across MRI, CT, and ultrasound modalities. Primarily descriptive and taxonomy-based. &
Does not include experimental comparison or preprocessing analysis. Our work performs standardized experiments on the CAMUS dataset using U-Net, Attention U-Net, and TransUNet. \\

Medical Image Segmentation with Domain Adaptation: A Survey \cite{li2023medical} &
Summarizes domain adaptation techniques for medical image segmentation across multiple organs and modalities. &
Focuses on cross-domain generalization rather than direct benchmarking. Our study maintains a single-domain (echocardiography) setting with reproducible training protocols. \\

A Review of Deep Learning-Based Methods for Medical Image Multi-Organ Segmentation \cite{fu2021review} &
Highlights architectures for multi-organ segmentation using CT and MRI data; focuses on cross-organ generalization and network design. &
Does not cover ultrasound modality or cardiac-specific tasks. Our study targets echocardiographic segmentation with controlled experimental evaluation. \\

Medical Image Analysis on Left Atrial LGE MRI for Atrial Fibrillation Studies: A Review \cite{li2022medical} &
Specialized review on left atrial segmentation and fibrosis quantification using LGE-MRI. &
Focuses on a single structure (left atrium) and imaging type (MRI). Our work extends to multi-structure (LV endocardium, myocardium, LA) segmentation on ultrasound. \\

Artificial Intelligence in Congenital Heart Disease: Current State and Prospects \cite{jone2022artificial} &
Discusses AI applications for congenital heart disease diagnosis and prognosis using multiple imaging modalities. &
Emphasizes disease detection and clinical applications, not technical segmentation evaluation. Our study focuses on segmentation accuracy and architecture-level insights. \\

This Work &
Controlled comparison of U-Net, Attention U-Net, and TransUNet architectures on the CAMUS dataset. Introduces SAM-based pseudo-labeling for semi-supervised training. &
Provides reproducible experimental benchmarking, analyzes preprocessing effects (NIfTI vs. PNG), and demonstrates quantitative gains from SAM-based label augmentation. \\
\hline
\end{tabular}
}
\label{tab:review_comparison}
\end{table*}

The remaining paper is organized as follows: Section \ref{cardiacbg} and \ref{carct} present a detailed review of cardiac imaging. Further, we discuss related work in Section \ref{RW}. Next, in Section \ref{benchmark} we discuss the benchmark setup. We then describe the results in section \ref{results}. Section \ref{discussion} presents the discussion and analysis. Finally, in section \ref{con}, we present the conclusion.

\section{Cardiac Imaging Background and Modalities}\label{cardiacbg}
\begin{figure*}[!ht]
     \centering
     \includegraphics[width=1.2\textwidth]{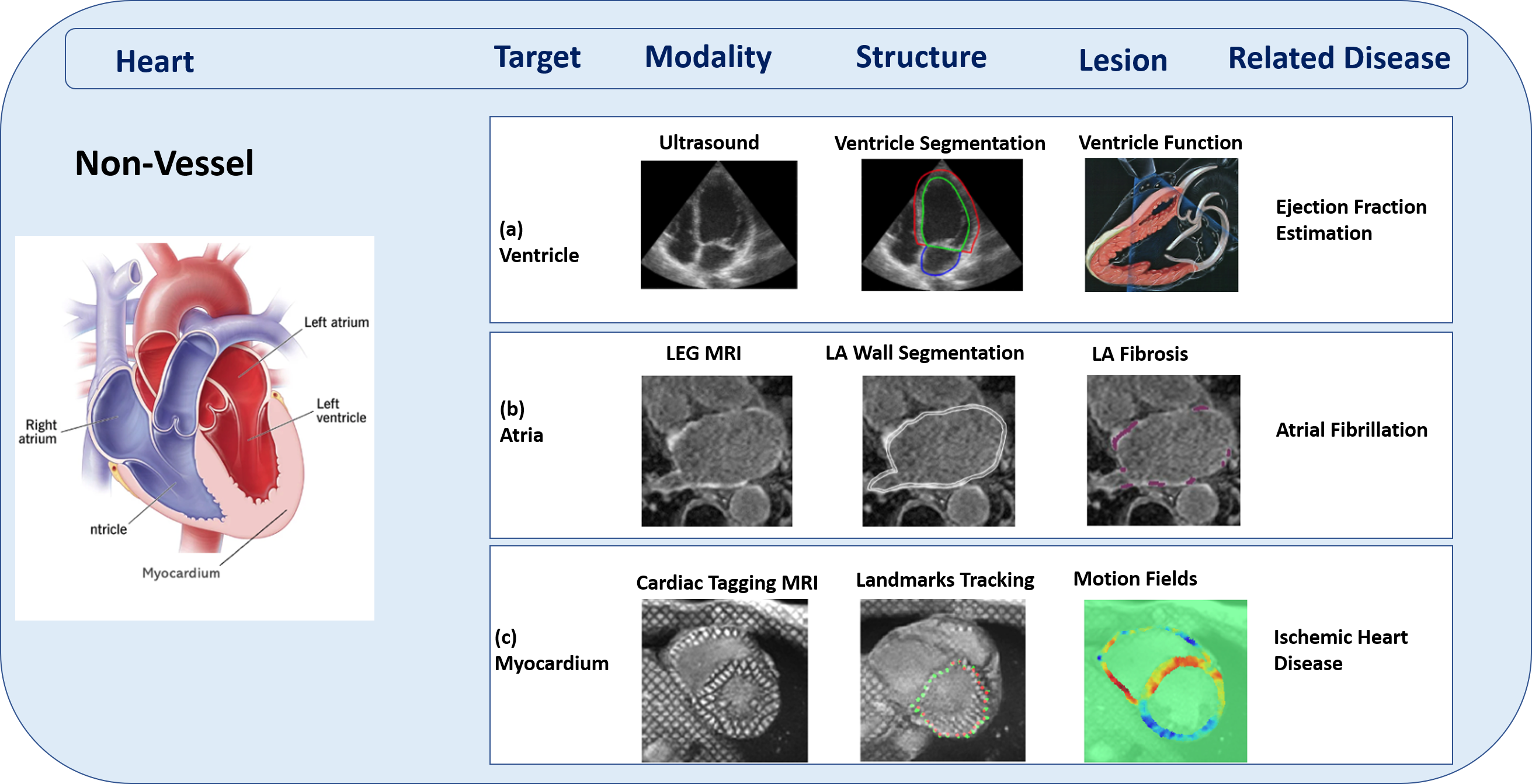}
     \caption{This figure presents a comprehensive overview of CVDs \cite{wang2024artificial}, categorized by the primary anatomical structures they affect and their corresponding physiological functions. It also outlines the key medical imaging modalities used for CVD diagnosis and analysis, emphasizing the growing contribution of AI in enhancing image-based assessment. The top section illustrates examples of non-vascular anatomical structures, including (a) ventricles \cite{leclerc2019deep} (b) atria \cite{li2022medical}, and (c) myocardium \cite{ye2021deeptag}. }
     \label{CVD}
 \end{figure*}

This section discusses the major imaging modalities commonly used in cardiac analysis as illustrated in Figure \ref{CVD}. Depending on the clinical objectives and the type of diagnostic information required, these techniques are applied individually or in combination to achieve an accurate and thorough evaluation of cardiac function and health.

\subsection{Cardiac Ultrasound}
Cardiac ultrasound imaging serves as a cornerstone in cardiovascular evaluation, utilizing sound waves to generate real-time visualizations of the heart \cite{zhou2021artificial,liu2023deep}. These images provide essential insights into the heart’s structure, including its size, shape, function, and blood flow dynamics \cite{aly2021cardiac}. Because echocardiography is non-invasive, free from ionizing radiation, and capable of delivering immediate results, it is widely favored as the first-line diagnostic tool for numerous cardiac conditions \cite{lu2023ultrafast}. It plays a vital role in detecting and monitoring disorders such as valvular heart disease, cardiomyopathies, and congenital defects \cite{jone2022artificial,ghorbani2020deep}. For vascular assessment, Intravascular Ultrasound is employed to obtain high-resolution images of the coronary arteries from within the vessel lumen \cite{xu2020fundamentals}. This technique uses a specialized catheter equipped with a miniature ultrasound transducer at its tip, which is guided through the coronary vessels to the targeted site. The resulting ultrasound images provide detailed visualization of the arterial wall structure and pathology \cite{de2002intravascular}.

\subsection{Cardiac X-ray Imaging}
Cardiac X-ray imaging, which includes both static radiographs and real-time fluoroscopy, is a diagnostic approach that employs X-rays to generate two-dimensional views of the heart and its vessels \cite{matsuoka2022deep}. It is frequently used in coronary angiography and proves especially valuable in emergency situations, offering rapid imaging with relatively low radiation exposure, making it suitable for most patients \cite{ccimen2016reconstruction}. This technique is highly effective for detecting CAD by directly visualizing the coronary vessels and guiding interventions such as angioplasty or stent placement to restore proper blood flow to the heart muscle \cite{yu2018coronary}. In cardiac X-ray imaging Digital Subtraction Angiography improves visualization by digitally eliminating background tissues, allowing clear focus on the blood vessels. When combined with contrast dye and rapid image capture, it produces high-resolution images that are essential for detecting vascular irregularities \cite{crabb2023deep}. 
Invasive Coronary Angiography is another specialized form of cardiac X-ray imaging primarily used during interventional procedures such as percutaneous coronary intervention (PCI). In PCI, a catheter is navigated through the coronary arteries to treat blockages, typically using angioplasty or stent placement. The X-ray imaging system provides real-time guidance for catheter positioning and enables continuous monitoring of the procedure’s progress \cite{niimi2022machine}.

\subsection{Cardiac Magnetic Resonance Imaging}
Cardiac Magnetic Resonance Imaging (MRI) is a non-invasive technique that uses strong magnetic fields and radiofrequency waves to generate detailed images of the heart \cite{saeed2015cardiac,bello2019deep}. It provides valuable information about cardiac anatomy, function, and blood flow, offering a comprehensive view of the heart’s structural and physiological characteristics. CINE MRI is a commonly used modality in cardiac imaging. Cine MRI records sequential images throughout the cardiac cycle, enabling real-time visualization of the heart’s motion. This technique facilitates detailed assessment of cardiac anatomy, ventricular volume, wall motion, and EF. It plays a key role in diagnosing conditions such as ventricular hypertrophy, myocardial infarction, and different types of valvular disease \cite{kustner2020cinenet,campello2021multi}. 

The next modality in cardiac MRI is cardiac tagging MRI (t-MRI), also known as myocaridal tagging. t-MRI provides a specialized method for analyzing myocardial motion and deformation. During the scan, a grid-like pattern, known as ``tags,” is applied to the heart muscle, allowing precise tracking of tissue movement. As the heart contracts and relaxes, these tags deform, offering detailed information about both global and regional myocardial function. Considered the gold standard for measuring regional strain and deformation, cardiac tagging plays a crucial role in diagnosing, managing, and studying heart diseases such as ischemic heart disease and dilated cardiomyopathy \cite{ye2021deeptag}.

Another form of cardiac MRI is T2-weighted MRI which is distinguished by its ability to enhance image contrast based on variations in tissue water content and physiological conditions. This characteristic makes it particularly effective for detecting myocardial abnormalities such as inflammation, edema, and ischemia. By clearly highlighting myocardial edema associated with inflammation or acute ischemic injury, T2-weighted MRI provides valuable diagnostic and therapeutic insights \cite{ren2022comparison}.

\section{Cardiac Computed Tomography}\label{carct}
Cardiac Computed Tomography (CT) uses advanced X-ray technology to generate high-resolution, three-dimensional cross-sectional images of the heart and nearby vessels. Although it involves a higher radiation dose than standard cardiac X-ray imaging, it provides much greater anatomical detail and diagnostic precision. Cardiac CT is especially useful for evaluating the coronary arteries, helping to detect narrowing or blockages that signal CAD \cite{garg2023role}. The following sections discuss the different Cardiac CT modalities and their specific uses in cardiac imaging \cite{gu2021cyclegan}.

Coronary CT Angiography (CTA) is a CT-based imaging technique that differentiates tissues by their density, enabling clear visualization of structures such as soft tissue, calcium deposits, fat, and air. This capability is particularly valuable for assessing the presence of calcium in the coronary arteries. When performed with a contrast agent, coronary CTA provides highly detailed images of the heart chambers, vessels, and coronary arteries, allowing accurate detection of non-calcified plaques \cite{lin2022deep}. It is extensively used to diagnose CAD, identify coronary anomalies, and assist in preoperative planning for bypass surgery. Moreover, it is useful for evaluating stent patency following implantation \cite{xu2023coronary}.

A Calcium-Scoring Heart Scan, also known as Coronary Artery Calcium scoring, is a specialized CT-based X-ray test that measures calcium deposits within the coronary arteries. These calcifications serve as early indicators of CAD, often appearing before symptoms develop. The resulting calcium score is used to assess the individual’s risk of CAD; higher scores correspond to an increased likelihood of future heart attacks and other cardiovascular events \cite{eng2021automated}.

Functional Cardiovascular CT is a non-invasive imaging approach used to evaluate both the structural and functional aspects of the heart and vascular system. It produces high-resolution cross-sectional images, allowing detailed assessment of myocardial perfusion and ventricular performance. This technique is particularly valuable in diagnosing and monitoring ischemic heart disease, cardiomyopathies, and heart failure \cite{peper2020functional}.

\subsection{Non-Vessel Structures}
\subsection{Ventricle}
The ventricles, which form the two lower chambers of the heart, are essential components of cardiac anatomy, as shown in Figure \ref{CVD}(a). The left ventricle (LV), the largest and most powerful chamber, receives oxygenated blood from the left atrium (LA) and pumps it into the systemic circulation, ensuring the delivery of oxygen and nutrients throughout the body. In contrast, the right ventricle (RV) collects oxygen-poor blood from the right atrium and propels it into the pulmonary circulation for oxygenation \cite{zhao2022deep,pantelidis2023deep}.

In this section, we discuss various ventricular disorders and examine how artificial intelligence is being applied to study and manage these conditions \cite{shoaib2023overview}. The focus is on the integration of AI with cardiology, emphasizing its growing role in improving the diagnosis, analysis, and treatment of ventricular diseases.

We examine several structural and functional abnormalities of the ventricles, each with distinct clinical features and implications \cite{olaisen2024automatic}. Abnormal heart rhythms often originate in the ventricles, ranging from mild cases such as premature ventricular contractions to severe conditions like ventricular tachycardia or fibrillation \cite{madan2022hybrid}. These arrhythmias disrupt normal cardiac pumping and may become life-threatening if untreated \cite{kolk2023machine}. Another example is the Ventricular Septal Defect, a congenital condition marked by an opening in the wall separating the ventricles. This defect causes oxygenated and deoxygenated blood to mix, reducing overall cardiac efficiency, with the impact depending on the size and location of the defect \cite{yang2023application}.

As illustrated in Figure \ref{CVD}(a), the EF serves as a vital indicator of the LV's pumping performance and helps identify patients at risk of heart dysfunctions such as heart failure \cite{kusunose2022standardize}. EF quantifies the percentage of blood ejected from the LV during each contraction \cite{liu2021deep}; under normal conditions, the ventricle expels more than half of its volume per beat, whereas a reduced EF indicates weakened cardiac function \cite{akerman2023automated}. A recent study proposed a Graph Neural Network–based model, EchoGNN, to estimate EF from echocardiography videos, achieving performance comparable to leading approaches while enhancing interpretability and addressing inter-observer variability \cite{mokhtari2022echognn}.

\subsection{Atria}
The atria, comprising the left and right chambers at the top of the heart, are responsible for receiving incoming blood, as illustrated in Figure \ref{CVD}(b). The right atrium gathers deoxygenated blood returning from the body, while the LA receives oxygen-rich blood from the lungs. Together, they play a vital role in directing blood into the ventricles. This section discusses common atrial disorders and explores how AI is being applied to their detection and diagnosis. We discuss several prevalent atrial disorders \cite{wang2022deep,ter2023juvenile}, with atrial fibrillation (AF) being the most common. AF is a widespread cardiac arrhythmia characterized by irregular and rapid electrical activity in the atria \cite{raicea2021giant,wang2022current}. Instead of contracting rhythmically, the atria quiver or fibrillate, producing an irregular heartbeat that can cause palpitations, shortness of breath, fatigue, and dizziness \cite{makynen2022wearable}. Importantly, AF increases the likelihood of stroke due to the formation of blood clots within the atria \cite{raghunath2021deep}.

To detect AF-related abnormalities, electrocardiography (ECG) and cardiac MRI are primarily used. ECG is preferred for real-time monitoring, while Late Gadolinium Enhancement (LGE) MRI provides detailed visualization of structural changes and fibrosis. DL models trained on large ECG datasets have demonstrated high accuracy in identifying AF patterns \cite{raghunath2021deep}, often performing on par with or better than clinical experts \cite{murat2021review}. As illustrated in Figure \ref{CVD}(a), AI-based AF analysis involves multiple stages, including LA cavity and wall segmentation, scar segmentation, quantification, and applications such as identifying ablation gaps in LGE MRI scans \cite{li2020atrial}.

Quantitative assessment of atrial abnormalities typically includes evaluating atrial volume, detecting fibrotic regions, and measuring atrial wall thickness. These parameters are extracted using image processing techniques and analyzed through shape and intensity-based metrics. Although these features are generally reliable, the variability of AF presentation poses challenges for standardization. For an in-depth overview of AI methodologies in AF research covering LA cavity, wall, and scar segmentation from LGE MRI, see the comprehensive review in \cite{li2022medical}.

\subsection{Myocardium}
The myocardium, shown in Figure \ref{CVD}(c), forms the muscular middle layer of the heart and plays a vital role in its overall function. Made up of specialized cardiac muscle cells known as cardiomyocytes, it is responsible for the rhythmic contractions and relaxations that enable blood circulation throughout the body. This tissue’s proper function is essential for sustaining heart performance and maintaining efficient blood flow \cite{rinaldi2022invasive}.

The myocardium is nourished by the coronary arteries, which deliver oxygen and vital nutrients to its cells. With every heartbeat, the myocardium contracts to push blood from the heart’s chambers into the circulatory system. However, various diseases can compromise this tissue, reducing its ability to contract effectively and support normal cardiac activity. The following section examines several common myocardial disorders and their clinical significance \cite{jafari2022automatic}, emphasizing the myocardium’s susceptibility and the importance of preserving its health for overall cardiovascular stability. Myocardial Infarction (Heart Attack) is a serious condition affecting the myocardium that occurs when blood flow to part of the heart muscle becomes blocked, usually due to a clot in the coronary arteries \cite{zheng2019explainable}. This obstruction can damage or destroy the affected tissue, leading to chest pain, breathing difficulty, and other severe complications \cite{cho2020artificial}. Recent advances in DL have made it possible to detect and quantify chronic myocardial infarction using non-contrast cardiac MRI, reducing the need for gadolinium-based contrast agents \cite{zhang2019deep}. In another approach, machine learning models have been developed to assess infarction risk by integrating cardiac troponin levels with clinical data \cite{doudesis2023machine}. Myocardial perfusion MRI, a noninvasive imaging method, is also widely used to identify ischemic heart disease with high precision. As shown in Figure \ref{CVD}(c), AI-driven models can generate detailed visualizations of cardiac blood flow, helping clinicians detect and assess areas with reduced perfusion and improving overall management of cardiac health.

\section{Background Studies}
\label{RW}

Accurate segmentation of cardiac structures from medical imaging modalities is an essential step for quantitative assessment in clinical cardiology. Over the years, numerous DL architectures have been proposed for automated segmentation, with a particular emphasis on echocardiography due to its non-invasive nature, low cost, and real-time imaging capability. Among these, U-Net, Attention U-Net, and TransUNet have become cornerstone architectures for medical image segmentation. This section presents a detailed overview of these models, their architectural characteristics, advantages, and relevance to cardiac image analysis, followed by a discussion of the CAMUS dataset and existing benchmarking limitations in the literature. Table \ref{tab:cardio_metrics}, illustrates multiple evaluation indices that are used to assess cardiac segmentation models. The selected metrics cover pixel-level accuracy (Dice, IoU), boundary conformity (HD, ASD), and clinical consistency (EF error, volume difference) to ensure a comprehensive evaluation across datasets such as CAMUS, ACDC, and EchoNet-Dynamic.

\begin{longtable}{p{4cm} p{1.7cm} >{\raggedright\arraybackslash}p{8.2cm}}
\caption{Cardiac image segmentation metrics, their units, and dataset-specific applications.}
\label{tab:cardio_metrics}\\

\toprule
\textbf{Metric} & \textbf{Unit} & \textbf{Application in Cardiac Segmentation} \\
\midrule
\endfirsthead

\toprule
\textbf{Metric} & \textbf{Unit} & \textbf{Application in Cardiac Segmentation} \\
\midrule
\endhead

\bottomrule
\endfoot

\bottomrule
\endlastfoot

Dice Similarity Coefficient (DSC) & 0--1 & Measures overlap accuracy between predicted and ground-truth regions (LV, RV, myocardium); widely reported for CAMUS and ACDC datasets. \\
Jaccard Index (IoU) & 0--1 & Evaluates intersection-over-union of segmented masks, complementary to Dice. \\
Hausdorff Distance (HD) & mm & Quantifies maximum boundary deviation; critical for M\&Ms dataset due to shape variability across scanners. \\
Average Surface Distance (ASD) & mm & Measures mean boundary distance; often used in ACDC leaderboard metrics. \\
Precision & 0--1 & Evaluates proportion of correctly predicted cardiac pixels among all predicted positives. \\
Recall (Sensitivity) & 0--1 & Quantifies ability to detect all true cardiac structures; key for LV cavity segmentation in CAMUS. \\
Specificity & 0--1 & Reflects the model’s ability to reject non-cardiac background pixels. \\
F1-score & 0--1 & Balances precision and recall for multi-class segmentation (LV, RV, myocardium). \\
Volume Difference (VD) & \% & Computes relative volume error between prediction and ground truth (e.g., LVEDV, LVESV). \\
Ejection Fraction Error (EFE) & \% & Evaluates error in predicted ejection fraction; important for clinical validation on EchoNet-Dynamic and CAMUS. \\
Myocardial Thickness Error (MTE) & mm & Measures discrepancy in myocardial wall thickness estimation (used in M\&Ms inter-scanner analysis). \\
Boundary F1-score (BF-score) & 0--1 & Evaluates contour precision at pixel level; sensitive to thin myocardial boundaries. \\
Normalized Surface Dice (NSD) & 0--1 & Computes boundary overlap within a tolerance (1--3 mm); increasingly adopted in M\&Ms-2 challenges. \\
Processing Time per Frame & seconds & Indicates computational efficiency; relevant for real-time echocardiographic analysis in CAMUS. \\
Memory Usage & MB & Measures resource efficiency, especially for 3D MRI datasets like ACDC. \\
Mean Absolute Error (MAE) & mm or \% & Quantifies mean difference in volume or boundary metrics across cases. \\
Clinical Score (LV Function Classification) & \% & Derived from segmentation-based functional indices to assess cardiac health categories (normal, mild, severe). \\

\end{longtable}

\subsection{U-Net Architecture}

The U-Net, first introduced by Ronneberger et al. \cite{ronneberger2015u}, is a fully convolutional neural network (FCN) specifically designed for biomedical image segmentation. Its defining feature is a symmetric encoder–decoder architecture with skip connections that bridge the corresponding layers in the downsampling and upsampling paths. The encoder progressively extracts semantic features through repeated convolution and pooling operations, while the decoder reconstructs the spatial resolution using transposed convolutions and concatenated feature maps from the encoder.

The skip connections enable the network to recover fine spatial details that might otherwise be lost during downsampling, allowing for accurate delineation of complex anatomical boundaries. U-Net’s simplicity, efficiency, and high segmentation accuracy have made it one of the most widely used architectures in medical imaging.

In the context of cardiac segmentation, U-Net has demonstrated high performance on multiple modalities, including MRI, CT, and echocardiography. Typical reported DSC for LV segmentation range between 0.90 and 0.96, depending on the dataset, image modality, and labeling consistency. For instance, studies on the CAMUS and ACDC datasets have achieved Dice scores above 0.93 for end-diastolic (ED) and end-systolic(ES) phases, validating U-Net as a strong baseline for cardiac image segmentation tasks.

\subsection{Attention U-Net}

While U-Net provides strong performance, its convolutional filters treat all spatial regions equally, which can be suboptimal when segmenting small or low-contrast anatomical structures, such as the apical regions or thin myocardium boundaries in echocardiographic views. To overcome this limitation, Attention U-Net (Oktay et al., \cite{oktay2018attention}) introduced attention gates (AGs) into the U-Net architecture. Attention gates are lightweight modules that learn to emphasize salient spatial regions while suppressing irrelevant or noisy activations in the feature maps. These gates are typically inserted along the skip connections, enabling the network to modulate information flow between the encoder and decoder dynamically. By assigning higher weights to task-relevant features (e.g., ventricular walls or endocardial boundaries), the model enhances its focus on critical structures and reduces segmentation errors in challenging regions.

In cardiac applications, Attention U-Net has been shown to improve segmentation precision for smaller chambers and boundary regions, particularly in echocardiographic data where speckle noise and signal dropout are common. Reported Dice improvements of 1–3\% over baseline U-Net models have been observed in several studies. Moreover, the attention mechanism enhances interpretability, as attention maps can visually indicate which regions the model prioritizes during prediction.

\subsection{TransUNet}

Recent advances in computer vision have highlighted the advantages of ViTs in modeling long-range dependencies and global context features that conventional CNNs struggle to capture due to their localized receptive fields. Building on this, TransUNet (Chen et al., \cite{chen2021transunet}) combines the local feature extraction capability of CNNs with the global reasoning power of Transformers, forming a hybrid CNN–ViT segmentation architecture. In TransUNet, the encoder is typically a convolutional backbone (e.g., ResNet or U-Net-like CNN) that extracts low-level spatial features. These feature maps are then flattened and passed through a Transformer encoder, which applies self-attention mechanisms to model spatial and contextual relationships across the entire image. The decoder mirrors the U-Net structure, progressively upsampling and fusing multi-scale representations to reconstruct fine segmentation boundaries.

This hybrid design enables TransUNet to retain local texture sensitivity from CNNs while incorporating global contextual information from Transformers. In cardiac segmentation tasks, this often translates into improved generalization, particularly for cases with anatomical variability or low-quality ultrasound frames. Several studies have reported Dice scores above 0.94–0.96 for TransUNet-based cardiac segmentation, surpassing traditional CNN baselines in challenging conditions.

\subsection{Model Complexity and Computational Efficiency}
To better contextualize the architectural differences between U-Net, Attention U-Net, and TransUNet, we report their parameter counts, computational cost, inference speed, and GPU memory consumption (see Table~\ref{tab:model_complexity}). U-Net is the most lightweight model, requiring only 7.8M parameters and 35 GFLOPs, which enables fast inference (~8 ms per 512×512 frame). Attention U-Net introduces attention gates to refine spatial focus, resulting in a moderate increase in parameters and FLOPs (45 GFLOPs), with a small reduction in throughput (~10 ms/frame). In contrast, TransUNet is substantially heavier, combining CNN and Transformer blocks, and therefore requires over 100M parameters and 140 GFLOPs. This leads to slower inference (~24 ms/frame) and the highest GPU memory footprint. 

\begin{table}[H]
\centering
\caption{Model complexity and inference speed for U-Net, Attention U-Net, and TransUNet on 512$\times$512 CAMUS images (RTX~3090).}
\resizebox{\textwidth}{!}{
\begin{tabular}{lcccccc}
\toprule
\textbf{Model} & \textbf{Params (M)} & \textbf{GFLOPs} & \textbf{Encoder} & \textbf{Attention} & \textbf{Time (ms)} & \textbf{Memory (GB)} \\
\midrule
U-Net          & 7.8   & 35  & CNN             & None     & 8.1  & 3.2 \\
Attention U-Net & 8.1  & 45  & CNN             & AG       & 10.4 & 3.8 \\
TransUNet      & 105   & 140 & CNN + ViT       & MHSA     & 23.6 & 8.9 \\
\bottomrule
\end{tabular}
}
\label{tab:model_complexity}
\end{table}

These measurements clarify the trade-off between accuracy and efficiency, highlighting that while TransUNet provides superior global context modeling, U-Net and Attention U-Net remain more practical for real-time or resource-constrained clinical settings.

\subsection{Review of Deep Learning Methods for Cardiac Segmentation}
\label{LR}

\begin{longtable}{p{3.0cm} p{3.0cm} p{4.5cm} p{4cm}}
\caption{Recent studies on cardiac segmentation.}
\label{recent}\\
\hline
\textbf{Author} & \textbf{Dataset} & \textbf{Performance} & \textbf{Architecture} \\
\hline
\endfirsthead

\caption*{Table \thetable\ (continued): Recent studies on cardiac segmentation.} \\
\hline
\textbf{Author} & \textbf{Dataset} & \textbf{Performance} & \textbf{Architecture} \\
\hline
\endhead

\hline \multicolumn{4}{r}{{Continued on next page}} \\ \hline
\endfoot

\hline
\endlastfoot

Leclerc et al. \cite{leclerc2019deep} & CAMUS Dataset &
CAMUS: LVEndo (ED: 0.939, ES: 0.916); LVEpi (ED: 0.954, ES: 0.945) &
U-Net 1, U-Net 2, ACNN, SHG, U-Net++ \\

Nathan et al. \cite{painchaud2020cardiac} &
ACDC, CAMUS Datasets &
ACDC: LV: 0.94–0.95, MYO: 0.89–0.91, RV: 0.89; CAMUS: LV endo: 0.90, LV epi: 0.88, LA: 0.86 &
Constrained Variational Autoencoder \\

Liu et al. \cite{liu2021deep} &
CAMUS, sub-EchoNet-Dynamic Datasets &
CAMUS: LV Endo (ED: 0.951, ES: 0.931); LV Epi (ED: 0.962, ES: 0.956);
sub-EchoNet: LV Endo (ED: 0.942, ES: 0.918); LV Epi (ED: 0.951, ES: 0.943) &
Deep Pyramidal Local Attention Network \\

Wu et al. \cite{wu2022semi} &
EchoNet-Dynamic, CAMUS Datasets &
EchoNet-Dynamic: LV Endo: 0.9287; CAMUS: LV Endo: 0.9379 &
Noise-Resilient Spatio-Temporal Semantic Calibration and Fusion Network \\

Feng et al. \cite{feng2023learning} &
Echocardiography, CAMUS Datasets &
Echocardiography: LA 1-shot: 0.8637, LV 1-shot: 0.8706; LA 5-shot: 0.8802, LV 5-shot: 0.8787; CAMUS: LA/LV (2C/4C): 0.4865 avg &
Registration and Prototype Network \\

Gilbert et al. \cite{gilbert2021generating} &
CAMUS, EchoNet-Dynamic, SiteA/SiteB, Pathological Datasets &
CAMUS: LV Endo: 0.91; EchoNet-Dynamic: LV Endo: 0.90; SiteA: LV Endo: 0.88; Pathological: LV Endo: 0.87 &
CycleGAN-based Framework \\

Nathan et al. \cite{painchaud2022echocardiography} &
CAMUS Dataset &
U-Net: 0.955, LU-Net: 0.952, ENet: 0.949, DeepLabv3: 0.947, CLAS: 0.956 &
U-Net, LU-Net, ENet, DeepLabv3, CLAS \\

Eman et al. \cite{alajrami2024active} &
Unity, CAMUS Datasets &
Unity: LV: 0.99; CAMUS: LV: 0.986 &
U-Net architecture with Monte Carlo Dropout \\

Stough et al. \cite{stough2020left} &
CAMUS Dataset &
LVendo (ED: 0.94, ES: 0.92); LVepi (ED: 0.96, ES: 0.95); LA (ED: 0.87, ES: 0.91) &
U-Net-based Encoder-Decoder with Adaptive Skip Connections \\

Li et al. \cite{li2022towards} &
ACDC, MSCMR Datasets &
ACDC: Dice correlation: 0.8879; MSCMR: 0.8785 &
U-Net \\

Khan et al. \cite{khan2025compositional} &
M\&Ms-2, CAMUS Datasets &
M\&Ms-2: LV: 0.9563, RV: 0.9193, MYO: 0.8761; CAMUS: LV: 0.9348, MYO: 0.8870, LA: 0.8990 &
Hierarchical U-Net-based Architecture \\

Tran et al. \cite{tran2016fully} &
Sunnybrook, LV Segmentation Challenge, RV Segmentation Challenge Datasets &
Sunnybrook: LV endo: 0.92, LV epi: 0.96; LV Challenge: 0.84; RV Challenge: RV endo: 0.84, RV epi: 0.86 &
Fully Convolutional Neural Network \\

Mahendra et al. \cite{khened2019fully} &
ACDC, LV-2011 Datasets &
ACDC: LV: 0.96, RV: 0.95, MYO: 0.89; LV-2011: 0.74 &
Fully Convolutional Multi-Scale Residual DenseNet \\

Zhuang et al.  \cite{zhuang2022cardiac} &
MS-CMRSeg 2019 Dataset &
MYO: 0.76, LV: 0.89, RV: 0.82 &
Supervised SK-U-Net \\

Wu and Zhuang \cite{wu2021unsupervised} &
ACDC, MSCMRSeg 2019, M\&Ms Datasets &
ACDC: LV: 0.950, MYO: 0.905, RV: 0.895; MSCMRSeg: LV: 0.913, RV: 0.872, MYO: 0.826; M\&Ms: 0.852 &
Variational Domain Adaptation Network \\

Oktay et al. \cite{oktay2017anatomically} &
UK Digital Heart, CETUS’14, ACDC Datasets &
UK Digital Heart: LV+MYO: 0.941; CETUS’14: LV cavity: 0.90; ACDC: 0.916 &
Anatomically Constrained Neural Network \\

Bernard et al. \cite{bernard2018deep} &
ACDC Dataset &
LV (ED: 0.950, ES: 0.930); RV (ED: 0.920, ES: 0.890); MYO (ED: 0.910, ES: 0.870) &
2D and 3D Variants \\

Hyuiyi et al. \cite{zhang2024convnextunet} &
ACDC, MM-WHS Datasets &
ACDC: LV: 0.952, MYO: 0.915, RV: 0.902 &
ConvNeXt U-Net \\

\end{longtable}

Leclerc et al. \cite{leclerc2019deep} presented a landmark study that introduced the CAMUS dataset, which remains one of the most comprehensive open-access echocardiographic resources. It includes 500 patients with both 2-chamber (2CH) and 4-chamber (4CH) ultrasound views, annotated for left ventricular endocardium (LVEndo), epicardium (LVEpi), and LA structures. The study benchmarked several encoder–decoder–based DL architectures, including U-Net, U-Net++, ACNN (Anatomically Constrained CNN), and Stacked Hourglass (SHG), comparing them to traditional methods like Structured Random Forest and BEASM. The results showed that U-Net (18M parameters) achieved the highest performance, with Dice scores of 0.939 (ED) and 0.916 (ES) for LVEndo, and 0.954 (ED) and 0.945 (ES) for LVEpi, surpassing classical segmentation algorithms and approaching inter-observer agreement levels. Clinical metrics such as EF and ES/diastolic volume correlations were also highly consistent with expert assessments. However, the model performance degraded on low-quality or artifact-heavy ultrasound images and smaller cardiac structures (e.g., LA). The dataset included only static ES and ED frames, lacking temporal information for dynamic motion analysis. Additionally, performance gains plateaued beyond 250 patients, suggesting limits in representational capacity under 2D-only training. Despite these constraints, the CAMUS dataset and U-Net–based framework established a strong foundation for reproducible echocardiographic segmentation research and inspired later developments in 3D and temporally consistent models.

Nathan et al. \cite{painchaud2020cardiac} proposed a post-processing framework that enforces strong anatomical guarantees in cardiac image segmentation. While CNNs such as U-Net or ACNN achieve high Dice scores on cardiac datasets, they often generate anatomically implausible shapes (e.g., cavities with holes or disconnected myocardium). To address this, the authors introduced a constrained variational autoencoder (cVAE) trained solely on valid cardiac shapes. The cVAE learns a smooth latent space representing anatomically correct structures, allowing any invalid segmentation to be projected and corrected toward the nearest valid shape. The framework is modality-independent and was evaluated on two standard datasets, such as ACDC (MRI short-axis) left and RV and myocardium. CAMUS (echocardiography long-axis) LV endocardium, epicardium, and LA. Their method eliminated anatomically invalid segmentations while maintaining accuracy comparable to inter-observer variability. Average Dice scores after correction remained around 0.90–0.95 for ACDC and 0.88–0.90 for CAMUS, with minimal change in Hausdorff distance or ejection-fraction error. This work is among the first to guarantee anatomically valid cardiac segmentations regardless of the input CNN’s output, without relying on explicit shape priors during training. It effectively bridges data-driven CNN segmentation and model-based anatomical constraints, improving reliability for clinical use.

Liu et al. \cite{liu2021deep} proposed a Deep Pyramid Local Attention Network (PLANet) to improve the segmentation of cardiac structures in 2D echocardiography, a task often challenged by low contrast, noise, and ambiguous boundaries. The model introduces two main innovations, such as Pyramid Local Attention (PLA), which focuses attention within localized regions of the feature map instead of globally, allowing the network to capture fine-grained structural details while reducing sensitivity to background noise. Label Coherence Learning (LCL), which enforces spatial consistency among neighboring pixels, leading to smoother and anatomically coherent contours. The method was evaluated on CAMUS and sub-EchoNet-Dynamic datasets, achieving high segmentation accuracy. These Dice scores surpass those of U-Net, U-Net++, and ACNN, showing better boundary preservation and anatomical reliability while maintaining real-time inference speed (61 fps). This work demonstrated that combining localized attention mechanisms with label-consistency learning effectively enhances both the precision and stability of cardiac segmentation in ultrasound images, making PLANet a strong and efficient benchmark model for clinical echocardiographic analysis.

Wu et al. \cite{wu2022semi} proposed a semi-supervised segmentation network for echocardiography videos, designed to address the challenges of speckle noise, limited annotations, and irregular cardiac motion in ultrasound imaging. Their architecture, termed Noise-Resilient Spatiotemporal Semantic Calibration and Fusion Network (NR-STSFNet), is built upon a mean-teacher semi-supervised framework that leverages both labeled and unlabeled frames for end-to-end training. The model consists of two key modules, such that Adaptive Spatiotemporal Semantic Calibration (ASSC), which aligns feature representations across consecutive frames at the feature level rather than at the pixel level, making the system robust to motion and noise. Bi-Directional Spatiotemporal Semantic Fusion (BSSF), which aggregates semantic information from adjacent frames in both temporal directions, enhancing temporal consistency and maintains structural continuity in cardiac motion. The obtained results outperform several existing CNN and Transformer-based methods, such as TransUNet, PLANet, and Joint-Motion, while maintaining strong anatomical coherence and smooth temporal transitions. However, its main drawbacks lie in limited cross-domain adaptability, dependency on temporal consistency, and higher computational overhead compared to simpler models.

Feng et al. \cite{feng2023learning} proposed RAPNet (Registration and Prototype Network), a two-branch few-shot segmentation framework that learns both what and where to segment in medical images. The model integrates a registration-based spatial branch to generate spatial priors and a segmentation branch with an attention-based fusion mechanism that combines spatial and prototype information for more accurate feature representation. Evaluated on echocardiography and CHAOS-MR datasets, RAPNet achieved high Dice scores up to 88.02\% on in-house echocardiography and 75.54\% on CHAOS MR outperforming several state-of-the-art few-shot segmentation models. Despite its strong results, the method has some limitations, such as it only captures a limited 3D spatial context (2.5D setup), depends heavily on accurate registration for generating reliable priors, can produce irregular edges in noisy ultrasound images, and introduces higher computational complexity due to its dual-branch design.

Gilbert et al. \cite{gilbert2021generating} proposed a novel CycleGAN-based framework for generating synthetic labeled echocardiography data using pre-existing 3D anatomical heart models to address the scarcity of annotated medical images. Their approach involves three main stages: generating pseudo-2D cardiac images from anatomical models, transforming them into realistic ultrasound-like images using CycleGAN, and training a U-Net segmentation network on this synthetic data. The method was evaluated on multiple real-world echocardiography datasets, including CAMUS, EchoNet-Dynamic, and two clinical datasets (SiteA and SiteB), achieving Dice scores of 87–91\% for LV segmentation comparable to models trained on real annotated data. This work demonstrated that synthetic data can effectively substitute for manual annotations in DL segmentation tasks. However, the study’s main limitations include reduced generalization to pathological cases (as the anatomical models were primarily from healthy hearts), computational demands associated with CycleGAN-based data generation, and the focus on static 2D images rather than dynamic or 3D cardiac motion.

Painchaud et al. \cite{painchaud2022echocardiography} introduced a temporal regularization framework in their paper to improve the temporal and anatomical smoothness of cardiac segmentation across the full cardiac cycle. Instead of analyzing only ED and ES frames, as done in conventional 2D CNN models like U-Net and DeepLabv3, the authors employed a Cardiac Shape Autoencoder (AR-VAE) that learns latent shape representations and enforces consistent transitions across consecutive frames. This post-processing module was applied to several baseline architectures, U-Net, LU-Net, ENet, DeepLabv3, and CLAS on an extended version of the CAMUS dataset containing annotations for 98 complete cardiac cycles. The proposed framework achieved small but consistent improvements in segmentation accuracy, raising the Dice score from 0.951 to 0.955 for U-Net and up to 0.956 for CLAS, while also reducing the Hausdorff distance by about 0.5 mm. However, the method has a few limitations: the temporal smoothing occasionally reduces accuracy at motion extremes (ED/ES), it increases computational cost due to optimization in latent space for each sequence, and it can oversmooth rapid cardiac motion, slightly suppressing true physiological variations. Overall, the study demonstrated that latent-space temporal regularization can significantly enhance temporal coherence in 2D+time echocardiographic segmentation with minimal loss of detail.

Eman et al. \cite{alajrami2024active} proposed an active learning framework for LV segmentation in echocardiography, aiming to reduce manual labeling while maintaining near-supervised performance. The method used a Modified U-Net architecture enhanced with Monte Carlo Dropout for Bayesian Active Learning, enabling uncertainty estimation and adaptive sample selection. The model was trained and evaluated on two datasets: the Unity dataset (1,224 studies, 2,800 frames from Imperial College Healthcare NHS Trust) and the CAMUS dataset (450 apical 4-chamber images). Using an Optimized Representativeness Sampling strategy combined with uncertainty-based selection, the model achieved 98.6\% of full-supervision accuracy on CAMUS and 99\% on Unity, while requiring only 20–30\% of the labeled data, effectively reducing annotation effort by up to 80\%. However, the study acknowledged several limitations: the approach’s performance depends on the diversity and feature distribution of the dataset, making generalization to unseen domains challenging; maintaining representative sample selection during active iterations requires fine-tuning; and the iterative querying process introduces additional computational overhead. Moreover, further testing on multi-view or pathological echocardiography datasets is needed to fully validate the method’s robustness across clinical variations.

Stough et al. \cite{stough2020left} introduced a U-Net–based CNNs for simultaneous segmentation of the LVEndo, epicardium (LVEpi), and LA in 2D echocardiography images. The model was trained and evaluated on the CAMUS dataset, which includes 450 patients with both apical 2-chamber (2CH) and 4-chamber (4CH) views, totaling 1,800 labeled images covering both ED and ES phases. The proposed CNN employed an encoder–decoder structure with additive skip connections and group normalization, optimized with cross-entropy (CE) loss and extensive data augmentation to improve generalization. The model achieved high segmentation performance, with Dice scores of 0.93 (ED) and 0.92 (ES) for LVEndo, 0.96 (ED) and 0.95 (ES) for LVEpi, and 0.87 (ED) and 0.91 (ES) for LA, matching or surpassing inter-observer agreement levels. These results demonstrate strong reliability for cardiac structure delineation and clinical parameter estimation, such as EF (mean absolute error 3.8\%). However, the model was trained exclusively on the CAMUS dataset, which limits its generalization to different scanners or imaging conditions. Additionally, since the approach was based on static 2D frames, it lacked temporal coherence across cardiac cycles and might not capture dynamic heart motion. The authors suggested that future work could integrate temporal modeling and domain adaptation to improve robustness and cross-domain applicability. 

Li et al. \cite{li2022towards} proposed a novel self-reflective framework for assessing the reliability of cardiac image segmentation at both the image and pixel levels. The method introduces a self-reflective reference generator based on a conditional GAN (CGAN) that reconstructs input images from predicted segmentations, enabling detection of segmentation errors by comparing reconstructed and original images. A difference investigator module further quantifies these discrepancies using a semantic class-aware compactness constraint, providing interpretable quality estimates for each structure. The framework was evaluated on two public cardiac MRI datasets i.e., ACDC and MSCMR, using U-Net as the baseline segmentation model, achieving strong per-class Dice correlations of 0.8879 (ACDC) and 0.8785 (MSCMR) and AUROC scores around 95\%, demonstrating its ability to accurately predict segmentation quality. However, reduced robustness to domain shifts across different scanners or imaging protocols, higher computational cost due to adversarial training, and limited interpretability for borderline or ambiguous segmentations. Despite these challenges, the study represents an important step toward reliable, automated quality control in cardiac image segmentation and paves the way for safer clinical deployment of DL models.

Khan et al. \cite{khan2025compositional} proposed a compositional segmentation framework that integrates image features and metadata to improve cardiac image segmentation across different modalities. The method employs a hierarchical U-Net–based architecture with two decoders: a super-segmentation decoder for global cardiac localization and a sub-segmentation decoder for fine-grained structure segmentation. A novel Cross-Modal Feature Integration (CMFI) module fuses metadata, such as scanner type, disease category, and patient demographics with visual features to enhance the model’s contextual understanding and generalization. The approach was evaluated on two public datasets, M\&Ms-2 (MRI) and CAMUS (ultrasound), achieving strong performance with average Dice scores of 91.72\% (M\&Ms-2) and 90.69\% (CAMUS), outperforming existing models like UNet, ResUNet, and TransUNet. However, the framework’s performance is highly dependent on metadata availability and quality, and incomplete metadata can lead to reduced accuracy. Additionally, the hierarchical multi-decoder design and CMFI module introduce greater computational complexity and potential overfitting to specific metadata distributions, limiting scalability across unseen domains.

Tran et al. \cite{tran2016fully}  introduced one of the first FCN architectures for automated cardiac segmentation in short-axis MRI. The proposed 15-layer FCN performs end-to-end pixel-wise segmentation of the LV and RV from 2D MRI slices without the need for manual feature extraction or patch-based processing. The model was trained and evaluated on three public datasets, Sunnybrook Cardiac Dataset (MICCAI 2009), LV Segmentation Challenge (LVSC, STACOM 2011), and RV Segmentation Challenge (RVSC, MICCAI 2012), achieving high segmentation accuracy with Dice scores of 0.92 (LV endocardium) and 0.96 (LV epicardium) on Sunnybrook, 0.84 (Dice equivalent) on LVSC, and 0.84–0.86 on RVSC. These results demonstrated that a simple, fully convolutional approach could achieve state-of-the-art accuracy while being computationally efficient. However, segmentation accuracy decreased in apical and basal slices due to weak boundary visibility, the model’s 2D architecture lacked temporal and 3D spatial context, and its performance depended on dataset consistency and scanner variability. 

Mahendra et al. \cite{khened2019fully} proposed a fully convolutional multi-scale residual DenseNet (DFCN) architecture for cardiac MRI segmentation and automated cardiac disease diagnosis. The method integrates DenseNet connectivity, Inception-style multi-scale feature extraction, and residual long-skip connections to achieve efficient gradient flow and strong feature reuse while reducing parameters and GPU memory. A dual loss function combining Dice and CE losses was introduced to handle class imbalance and enhance boundary accuracy. The model was evaluated on three publicly available datasets, ACDC-2017, LV-2011, and Kaggle Data Science Bowl 2015, achieving high segmentation performance with Dice scores of 0.96 (LV endocardium), 0.95 (RV endocardium), and 0.89 (myocardium) on ACDC, and a Jaccard index of 0.74 on LV-2011, the best reported among fully automated algorithms at the time. The proposed approach also reached 100\% accuracy in automated cardiac disease diagnosis on the ACDC challenge test set. Despite strong results, the approach operates on 2D slices, which limits temporal and inter-slice spatial consistency. The reliance on ROI cropping and handcrafted features for disease classification makes the framework partially dependent on pre-processing. Moreover, its evaluation was limited to specific challenge datasets, which may not fully capture the variability of real-world clinical data. Future extensions could involve 3D temporal modeling and domain adaptation for multi-center robustness.

Zhuang et al. \cite{zhuang2022cardiac} presented a comprehensive benchmark study which evaluated multiple DL methods for myocardium, LV, and RV segmentation in LGE cardiac MRI. The study introduced the MS-CMRSeg 2019 dataset, containing 45 subjects with three CMR sequences, LGE, T2-weighted, and bSSFP, acquired using Philips Achieva 1.5 T scanners. Only five labeled LGE scans were provided for training, while 40 unlabeled LGE scans were used for testing, making this a highly challenging benchmark for domain adaptation and multi-sequence fusion. Nine representative algorithms were evaluated, including U-Net, Res-UNet, CycleGAN + U-Net, and MAS + AH-Net. The best-performing supervised model (SCU) achieved Dice scores of 0.926 (LV), 0.890 (RV), and 0.843 (myocardium), while the top unsupervised domain adaptation model (ICL) obtained 0.919 (LV), 0.875 (RV), and 0.826 (myocardium), demonstrating competitive performance between supervised and unsupervised methods. However, the drawbacks are, the segmentation accuracy significantly degraded in apical and basal slices, mainly due to poor tissue contrast in LGE images. Moreover, the limited labeled dataset (n = 5) constrained model generalization, and Hausdorff distances remained higher than inter-observer variability, indicating contour inconsistencies. The benchmark also relied on 2D slice-wise methods, lacking full 3D or temporal modeling, which limits spatial continuity across slices. 

Wu and Zhuang \cite{wu2021unsupervised} proposed an advanced unsupervised domain adaptation (UDA) framework called the Variational Domain Adaptation Network (VDAN) for cardiac MRI segmentation. The method addresses the performance gap between labeled and unlabeled cardiac MRI datasets by combining variational inference and adversarial learning to align latent feature distributions between source and target domains. Using ACDC as the labeled source domain and MSCMRSeg 2019 (LGE MRI) as the unlabeled target domain, the model effectively transfers segmentation knowledge without requiring manual annotations in the target data. VDAN achieved high segmentation accuracy, reporting Dice scores of 0.950 (LV), 0.905 (MYO), and 0.895 (RV) on the ACDC dataset, and an average Dice of 0.870 on MSCMRSeg 2019 (LV: 0.913, RV: 0.872, MYO: 0.826). The model also maintained good cross-domain generalization on the M\&Ms dataset with an average Dice of 0.852, outperforming previous UDA approaches such as CyCADA and SIFA. However, the training process is computationally expensive due to the dual use of variational approximation and adversarial learning. Additionally, the model’s performance depends on the similarity between source and target domains, and since it operates on 2D slices, it lacks spatial and temporal continuity across volumes. 

Oktay et al. \cite{oktay2017anatomically} introduced Anatomically Constrained Neural Networks (ACNNs), a DL framework that integrates anatomical shape priors into CNN training for cardiac image enhancement and segmentation. The method employs a regularization network (autoencoder) to learn valid anatomical representations, which are then used to constrain the segmentation predictions of the main CNN. This approach ensures that the predicted cardiac structures, such as the LV and myocardium (MYO) are not only accurate at the pixel level but also anatomically plausible. The model was evaluated on several datasets, including UK Digital Heart cine-MRI, CETUS’14 3D ultrasound, and ACDC 2017 cine-MRI, achieving average Dice scores of 0.941 (MRI) and 0.909 (ultrasound), outperforming standard 2D-FCN and 3D-UNet baselines. Despite its accuracy and shape-consistent predictions, the ACNN framework has high computational cost due to the dual-network (segmentation + shape prior) structure and requires pretraining on large, well-annotated datasets to learn reliable anatomical priors. Moreover, it primarily handles static 3D data, without explicitly modeling temporal information, which restricts its applicability in dynamic cardiac analysis such as full cardiac cycle segmentation or motion tracking. In summary, ACNN pioneered the integration of anatomical priors in DL for cardiac segmentation, significantly improving accuracy and shape realism, but at the cost of higher training complexity and limited temporal modeling.

Bernard et al. \cite{bernard2018deep} conducted a large-scale evaluation of DL methods for automatic cardiac MRI segmentation and diagnosis. The study was based on the ACDC dataset, which includes 150 cine-MRI cases divided into five diagnostic categories: normal, dilated cardiomyopathy (DCM), hypertrophic cardiomyopathy (HCM), myocardial infarction (MINF), and abnormal right ventricle. Each case contains expert-labeled LV, right ventricle (RV), and myocardium (MYO) masks at ED and ES phases. Several CNN-based architectures, primarily 2D and 3D U-Net variants, were benchmarked for segmentation and classification. The top-performing model \cite{isensee2017automatic} achieved Dice scores of 0.95 (LV), 0.91 (MYO), and 0.92 (RV), demonstrating near-human accuracy, with a mean Dice of 0.915 across structures. The study revealed several open challenges: segmentation accuracy degraded in apical and basal slices due to weak boundary contrast; performance for RV and MYO was less consistent than for LV; and the lack of temporal information in ED–ES pairs limited dynamic cardiac analysis. Furthermore, even though Dice and Hausdorff metrics showed high overlap, they did not always reflect anatomical plausibility. This benchmark demonstrated that DL, especially U-Net–based models, can achieve expert-level accuracy on the ACDC dataset, but challenges remain in anatomical consistency, edge cases, and temporal modeling, suggesting that the problem is not yet fully solved.

Huiyi and Zemin \cite{zhang2024convnextunet} introduce a hybrid DL architecture that combines the ConvNeXt backbone with a U-Net decoder enhanced by a small-region attention (SRA) mechanism to better capture fine structural details in cardiac MRI images. The model was evaluated on two benchmark datasets: ACDC (Automated Cardiac Diagnosis Challenge) and MM-WHS (Multi-Modality Whole Heart Segmentation) to test both accuracy and cross-modality generalization. On the ACDC dataset, ConvNextUNet achieved Dice scores of 0.956 for the LV, 0.918 for the myocardium (MYO), and 0.905 for the RV, resulting in an overall mean Dice of approximately 0.926, outperforming traditional U-Net and ResUNet models. The study highlights that the integration of ConvNeXt’s hierarchical feature extraction with U-Net’s localization capability allows for superior segmentation, particularly in small and low-contrast cardiac regions. However, the model incurs high computational cost due to the complex ConvNeXt backbone and attention modules, struggles slightly with apical and basal slices where boundaries are less distinct, and relies solely on 2D slice-based segmentation, which limits its ability to capture 3D or temporal cardiac motion information. Overall, ConvNextUNet demonstrates robust segmentation accuracy and improved attention to small regions but requires further optimization for efficiency and spatiotemporal modeling.

\subsection{Public Cardiac Imaging Datasets}\label{pub}
In recent years, several large-scale public image datasets have been introduced for cardiovascular research, greatly promoting collaboration and innovation in the field. These publicly available datasets are essential for advancing AI in cardiac imaging, as they provide diverse and well-annotated data needed for algorithm development and evaluation. By enabling shared benchmarks and collaborative experimentation, they help unify research efforts and accelerate progress. Moreover, the open accessibility of such datasets strengthens model training and validation, supporting the smooth integration of AI tools into clinical workflows and ultimately contributing to improved patient care. Table \ref{publicdatasets} classifies the available public datasets according to their primary focus areas and imaging modalities. The following sections provide a detailed overview of selected datasets, highlighting their structure, applications, and contributions to CVD research.

\begin{table}[!ht]
\centering
\caption{Publicly available cardiac image datasets on cardiac image analysis.}
\label{publicdatasets}
\begin{tabular}{
p{3.2cm}
p{2.2cm}
p{2.2cm}
p{2.5cm}
p{3.0cm}
}
\toprule
\textbf{Dataset} & \textbf{Modality} & \textbf{Task} & \textbf{Target} & \textbf{Disease Type} \\
\midrule
CAMUS \cite{leclerc2019deep} & 2D Ultrasound & Segmentation & LV, LA & Ejection fraction \\
ACDC (MICCAI) \cite{acdc2017} & Cine MRI & Segmentation & LV, RV, MYO & Myocardial infarction \\
MSC-MR \cite{zhuang2022cardiac} & CMR & Segmentation & LV, MYO, RV & Cardiomyopathy \\
M\&Ms-2 \cite{khan2025compositional} & Cine MRI & Segmentation & LV, RV, MYO & Cardiomyopathy \\
Sunnybrook Cardiac Data \cite{sunnybrook} & Cine MRI & Segmentation & LV, MYO & Hypertrophy, heart failure \\
LV Segmentation Challenge \cite{suinesiaputra2011left} & Cine MRI & Segmentation & LV, MYO & Coronary artery disease \\
RV Segmentation Challenge \cite{petitjean2015right} & Cine MRI & Segmentation & RV & Myocarditis \\
EchoNet-Dynamic \cite{ouyang2020video} & Echocardiography video & Segmentation & LV & Cardiomyopathy, heart failure \\
\bottomrule
\end{tabular}
\end{table}


\subsection{CAMUS Dataset}

The CAMUS dataset \cite{leclerc2019deep} is one of the most widely used benchmarks for cardiac ultrasound segmentation. It consists of 2-chamber (2CH) and 4-chamber (4CH) echocardiographic views acquired from 500 patients, annotated by expert cardiologists. The dataset includes ED and ES frames for each patient, with manual segmentations of the LV endocardium, LV epicardium, and left atrium (LA). Each patient record contains approximately four key images per view (ED/ES × 2CH/4CH), leading to a total of 2,000 annotated frames. The CAMUS dataset is split into training, validation, and test sets, ensuring patient-level separation to prevent data leakage. Its diverse patient population, variability in acquisition quality, and inclusion of both normal and pathological cases make it an ideal testbed for evaluating segmentation robustness.

To facilitate more extensive training, recent studies have also leveraged full-sequence data (cine loops) or frame-wise PNG/NIfTI exports, allowing networks to learn from unlabeled frames through self-supervised or semi-supervised approaches. This has enabled training on tens of thousands of frames, improving generalization and stability.

\subsection{Existing Benchmarks and Limitations}

While numerous studies have reported strong segmentation results on CAMUS, the lack of consistency in experimental protocols poses a major barrier to objective comparison across models. Many prior works differ in one or more of the following aspects, such as Preprocessing pipelines, some normalize intensities in [0,1], others z-score standardize, or apply histogram equalization. Differences in contrast enhancement, resizing, and cropping can alter model behavior. Data formats: Some use native NIfTI volumes with full dynamic range, while others convert to 8-bit or 16-bit PNGs, potentially losing grayscale fidelity. Training/validation splits: Patient overlaps or frame-level splits occasionally occur, leading to optimistic results. Loss functions and evaluation: Variations in CE, Dice, or hybrid losses complicate fair comparison, as do inconsistent metrics (e.g., per-frame vs. per-patient Dice). Due to these inconsistencies, most publications report isolated results, making it unclear whether observed improvements stem from architectural advances or from differences in preprocessing and hyperparameters.

\begin{figure*}[!ht]
     \centering
     \includegraphics[width=1\textwidth]{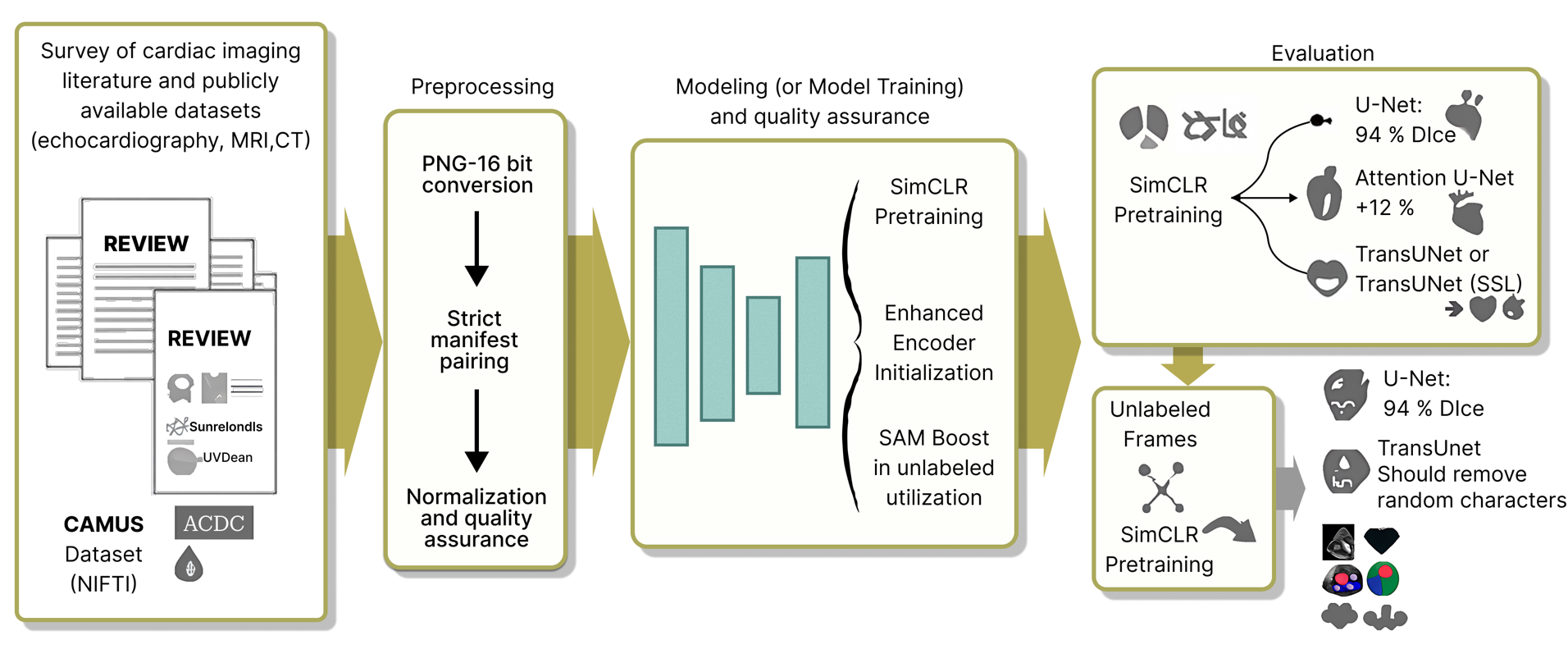}
     \caption{Unified workflow for cardiac ultrasound segmentation combining literature review, strict preprocessing, model training, and evaluation. The left block summarizes the survey of cardiac imaging literature and datasets (including CAMUS and ACDC). The preprocessing stage applies PNG-16-bit conversion, strict manifest pairing, normalization, and quality checks. The modeling stage integrates U-Net, Attention U-Net, and TransUNet architectures, supported by SimCLR pretraining and SAM-based pseudo-labeling to improve encoder initialization and make use of unlabeled frames. The evaluation module compares model performance across architectures, quantifies gains from SSL, and highlights improvements in segmentation accuracy and anatomical consistency. The workflow emphasizes how preprocessing fidelity, SAM, and SSL improve reproducibility and strengthen downstream segmentation outcomes. }
     \label{workflow}
 \end{figure*}

To address these gaps, this study performs a comprehensive, controlled comparison of U-Net, Attention U-Net, and TransUNet on the CAMUS dataset using identical preprocessing, data splits, loss functions, and evaluation metrics. The experiments also assess the effect of data representation (NIfTI vs. PNG-16bit) and self-supervised pretraining (SSL) on model performance and generalization. This unified analysis provides clear insights into the trade-offs among architectures and establishes reproducible baselines for future echocardiographic segmentation research (see Fig. \ref{workflow}).

\section{Benchmark Setup: Data, Models, and Training}\label{benchmark}

This section describes the dataset, preprocessing pipelines, network architectures, training setup, and evaluation procedures used for a controlled comparison between U-Net, Attention U-Net, and TransUNet on the CAMUS dataset. All experiments were conducted under standardized conditions to ensure reproducibility and fair comparison.

\subsection{Cardiac Ultrasound (Echocardiography)}
Cardiac ultrasound, or echocardiography, is a key technique in cardiovascular evaluation that uses high-frequency sound waves to generate real-time images of the heart \cite{zhou2021artificial,liu2023deep}. These images provide essential information about the heart’s structure, including its size, shape, functional performance, and blood flow dynamics \cite{aly2021cardiac}. Because echocardiography is non-invasive, free from ionizing radiation, and capable of producing immediate diagnostic results, it is widely used as a first-line imaging method for a variety of cardiac conditions \cite{lu2023ultrafast}. It plays an important role in detecting and managing disorders such as valvular defects, cardiomyopathies, and congenital abnormalities \cite{ghorbani2020deep,jone2022artificial}.

For vascular assessment, Intravascular Ultrasound serves as a specialized imaging technique that enables direct visualization of the coronary arteries from within \cite{xu2020fundamentals}. This method uses a catheter equipped with a miniature ultrasound transducer at its tip, which is carefully inserted through the coronary arteries to the target site. The ultrasound waves emitted from the probe capture detailed cross-sectional images of the arterial walls, allowing clinicians to evaluate plaque buildup, vessel morphology, and other structural features of the coronary vasculature \cite{de2002intravascular}.

\subsection{Dataset}
The CAMUS dataset was chosen because it is a clinically validated and widely used echocardiography benchmark with limited expert annotations, challenging image characteristics, and abundant unlabeled frames, making it well-suited for evaluating SAM-assisted pseudo-label generation and semi-supervised segmentation.
\subsubsection{CAMUS Dataset Overview}
The CAMUS dataset \cite{leclerc2019deep} is a benchmark echocardiography dataset containing apical two-chamber (2CH) and four-chamber (4CH) views from 500 patients. Each patient record includes both ED and ES frames. The ground truth segmentation masks are annotated by expert cardiologists for three cardiac structures, such as, LV Endocardium, LV Myocardium (Epicardium), and  LA. Each patient contributes four labeled frames (2 views $\times$ 2 phases), resulting in approximately 2,000 annotated images. In addition, the dataset provides cine-loop sequences in NIfTI (.nii.gz) format, offering thousands of unlabeled frames for semi- or self-supervised learning.

\subsubsection{NIfTI Volumes vs. PNG Converted Frames}
The dataset was evaluated under multiple preprocessing representations to investigate how data format and pairing strategies affect model performance. Four key pipelines were used: 

\paragraph{(i) Direct NIfTI Loading (Baseline Reference)}
The original volumetric NIfTI files were used directly, preserving full dynamic range and spatial resolution. Each frame was normalized on-the-fly within its native intensity distribution. This configuration serves as the reference baseline.

\paragraph{(ii) PNG 16-bit Conversion}
All frames were exported to 16-bit grayscale PNGs to retain intensity depth comparable to the NIfTI source while allowing efficient disk access. Robust z-score normalization and percentile clipping (0.5–99.5\%) were applied to mitigate outlier influence.

\paragraph{(iii) Strict PNG Dataset with Manifest Pairing}
To ensure accurate image–mask correspondence, a manifest file (\texttt{manifest.csv}) was generated, mapping each PNG image to its associated segmentation mask. Filenames were normalized by removing suffixes (e.g., \_mask, \_gt, \_seg), ensuring one-to-one pairing consistency and preventing mismatches.

\paragraph{(iv) All-Slice vs. Middle-Slice Strategies}
Two dataset variants were created, such as, Middle-slice strategy, where only the central frame per cine sequence was selected. All-slice strategy, where all frames from each volume were used, increasing the dataset to over 21,000 labeled frame pairs and 43,000 unlabeled frames. This larger set was utilized for self-supervised pretraining.

\subsection{Segment Anything Model Based Pseudo-Label Generation}\label{saam}

To expand the labeled training data for cardiac structure segmentation, we employed the SAM from Meta AI as a self-supervised pseudo-label generator. The goal was to automatically segment endocardial and myocardial regions from unlabeled echocardiographic frames in the CAMUS dataset and subsequently integrate these pseudo-labels for semi-supervised training of U-Net, Attention U-Net, and TransUNet models.

\subsubsection{Data Preparation}
The CAMUS dataset was first preprocessed into 16-bit PNG frames for both 2CH and 4CH sequences. For each patient, ED and ES frames were extracted from the NIfTI volumes and stored in the directory. The GT masks corresponding to the LV endocardium, LV myocardium, and LA were available for 500 patients, while all remaining frames were treated as unlabeled data.

\subsubsection{Automatic Mask Generation}
SAM was applied in automatic mask generation (auto) mode using the ViT-H checkpoint. For each echocardiographic frame: (i) The RGB image was resized to 512$\times$512 pixels and normalized. (ii) SAM generated multiple candidate regions per image. (iii) Each region was saved as a binary mask along with metadata fields such as predicted\_iou, area, and stability\_score. The output masks were stored as JSON files in the sam\_auto\_out/ directory, with one JSON file per frame.

\subsubsection{Post-Processing and Filtering}
To refine the raw SAM outputs, a custom post-processing script parsed the JSON mask files and converted valid masks into single-channel label maps. The filtering process was designed to retain only reliable regions according to the following criteria, if pred\_iou $\geq$ 0.7, or area $\geq$ 200 pixels, or top-3 highest scoring masks retained per frame. The resulting binary masks were merged into integer label maps 0=background, 1-3=structures. Each frame produced two output files:
(integer label map) and (RGB visualization).
These label maps were saved in the directory.
A manifest file was automatically built to index the image label pairs, resulting in a total of 68,447 pseudo-labeled frames.

\subsubsection{Semi-Supervised Integration}
The pseudo-labels were combined with the manually annotated data in a semi-supervised training framework.
GT samples were assigned a weight of 1.0, whereas SAM pseudo-labels were down-weighted to 0.5 to mitigate noise.
The combined manifest enabled training of the U-Net family using both real and pseudo annotations.
The training used a composite loss function combining CE and Dice losses, optimized with the Adam optimizer and gradient clipping (1.0).

\subsection{Results on SAM-Based Segmentation}

\subsubsection{Quantitative Evaluation}
When evaluating the pseudo-labels directly against the available GT masks on the CAMUS validation subset, the following Dice scores were obtained:

\begin{table}[H]
\centering
\caption{Performance of SAM pseudo-labels on CAMUS validation subset.}
\begin{tabular}{lcc}
\hline
\textbf{Structure} & \textbf{Dice Score} \\
\hline
LV Endocardium (c1) & \textbf{0.9988} \\
LV Myocardium (c2)  & 0.6694 \\
Left Atrium (c3)    & 0.6238 \\
\hline
\end{tabular}
\end{table}

The results show that SAM achieved near-perfect boundary detection for the LV endocardial cavity, whereas performance decreased for the myocardial and atrial regions due to their low contrast and shape variability in ultrasound images.

\subsubsection{Qualitative Analysis}
Visual inspection confirmed that SAM consistently captured the LV cavity with smooth and anatomically consistent contours. However, the myocardial boundaries were occasionally incomplete, and the atrial region exhibited inconsistent delineation due to acoustic shadowing and limited contrast. Nonetheless, the generated masks were sufficiently accurate to serve as pseudo-supervision for semi-supervised training.

\subsubsection{Discussion}
This experiment highlights that a foundation segmentation model, pre-trained on large-scale natural image datasets, can generalize effectively to echocardiographic ultrasound data without task-specific fine-tuning. Although myocardial and atrial delineations remain imperfect, SAM-generated pseudo-labels substantially increased the effective size of the labeled dataset and improved the generalization capability of the downstream CNN and Transformer-based models.


\begin{figure}[!t]
    \centering
    \includegraphics[width=\textwidth]{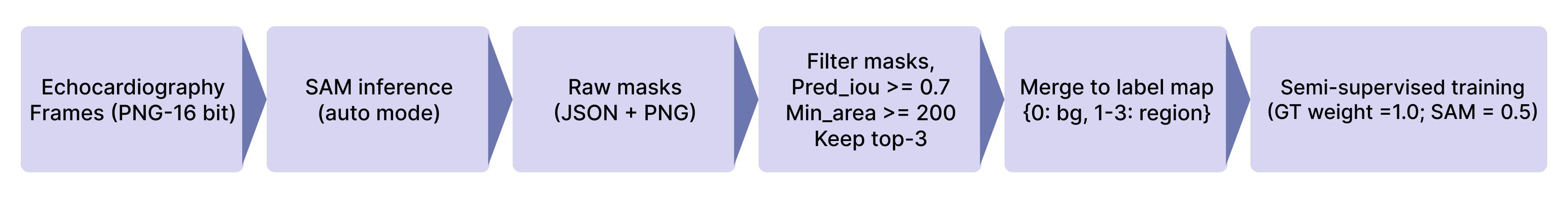}
    \caption{Pipeline for SAM-based pseudo-label generation and integration into the semi-supervised training framework. 
    The process begins with echocardiographic frames in PNG-16~bit format, which are passed through SAM in automatic mode to produce raw segmentation masks. 
    These masks are stored as JSON and PNG files, filtered by predicted IoU ($\geq 0.7$), minimum area ($\geq 200$~pixels), and top 3 mask selection. 
    The remaining masks are merged into structured label maps with three cardiac regions (endocardium, myocardium, and atrium) and incorporated into a semi-supervised training setup using GT and SAM pseudo-labels weighted at 1.0 and 0.5, respectively.}
    \label{fig:sam_pipeline}
\end{figure}

The overall pseudo-label generation and integration workflow is illustrated in Figure~\ref{fig:sam_pipeline}. Echocardiographic frames were first exported to 16-bit PNG format and processed by SAM in automatic segmentation mode. SAM produced multiple candidate masks per frame, stored as JSON and PNG pairs. A filtering procedure retained only the top-3 masks with predicted IoU $\geq 0.7$ and minimum region area $\geq 200$ pixels, ensuring anatomical plausibility and removing spurious noise. The selected regions were merged into a three-class label map, representing the LVEndo, myocardium, and LA, and subsequently incorporated into semi-supervised U-Net training, where manual GT labels and SAM pseudo-labels were weighted 1.0 and 0.5, respectively.

To exploit the unlabeled echocardiographic frames available in the CAMUS dataset, the SAM was adopted for automatic mask generation. 
Panel~(a) shows a raw echocardiographic input frame, while panel~(b) presents the corresponding SAM automatic segmentation overlay. 
Panel~(c) displays the coarse mask output containing multiple overlapping candidate regions, and panel~(d) depicts the final refined label map obtained after applying filtering criteria based on the top 3 masks with predicted IoU $\geq$~0.7 and minimum area $\geq$~200~pixels. 
These refined pseudo-labels represent three anatomical structures, the LVEndo, LV myocardium, and LA, colored as red, green, and blue, respectively.

The resulting pseudo-labels were combined with manually annotated ground-truth (GT) masks to form a semi-supervised training set. 
This hybrid dataset enabled effective learning from both expert-annotated and automatically labeled frames, thereby improving the model's generalization across varying cardiac phases and views. 

\subsection{Models}

\subsubsection{Baseline U-Net}
\begin{figure*}[!ht]
     \centering
     \includegraphics[width=1\textwidth]{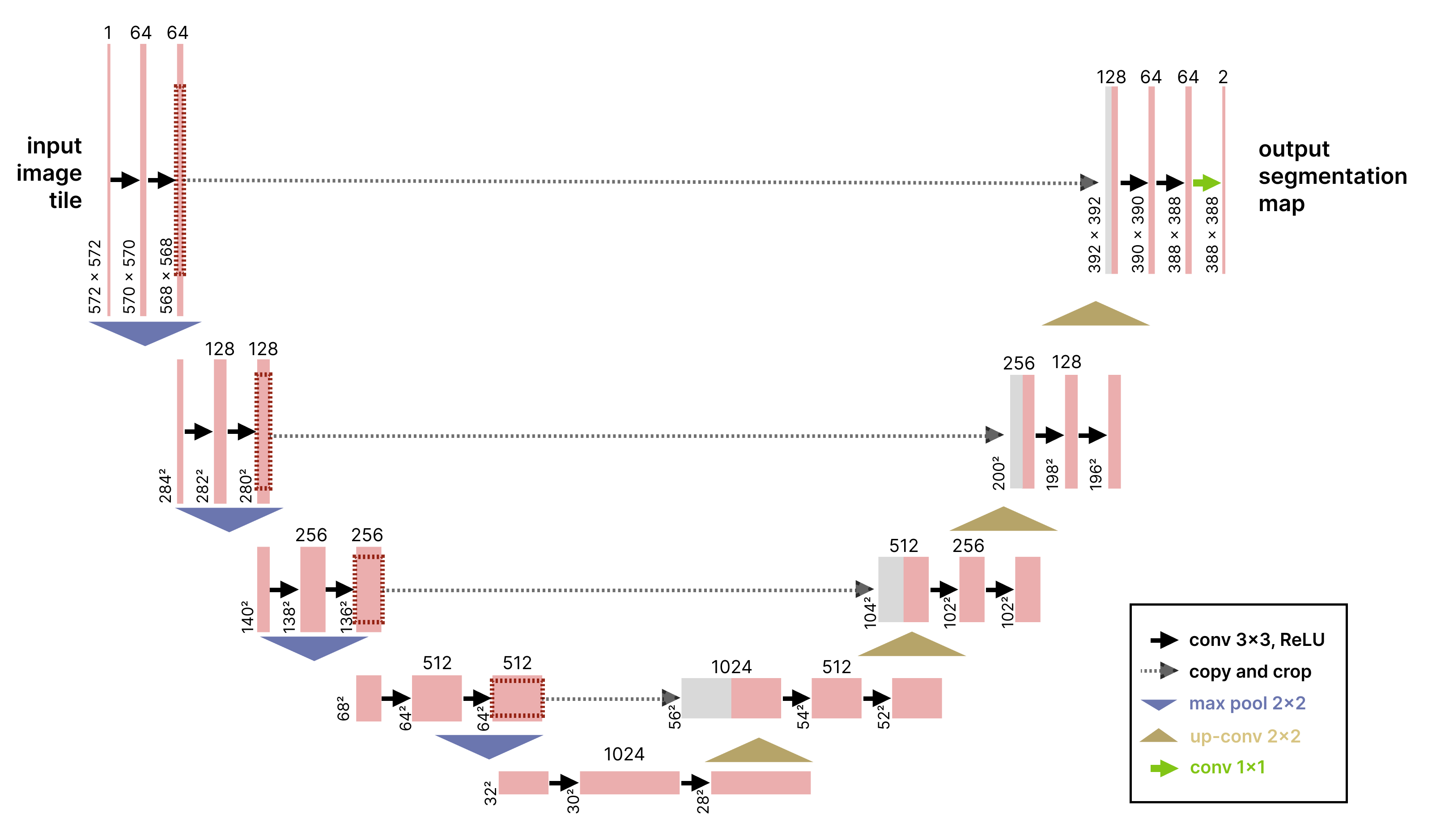}
     \caption{ Architectural diagram of the original U-Net model \cite{ronneberger2015u}. The network follows a symmetric encoder–decoder structure with skip connections that copy and crop feature maps from the contracting path to the expanding path, preserving spatial detail. Each encoder stage consists of two $3\times3$ convolution layers with ReLU activation followed by $2\times2$ max-pooling for downsampling. The decoder mirrors this structure using $2\times2$ up-convolutions to progressively recover spatial resolution. The final $1\times1$ convolution maps feature channels to the segmentation output. This architecture enables precise localization and has become the foundational backbone for biomedical image segmentation.}
     \label{unet}
 \end{figure*}

As shown in Fig. \ref{unet} The baseline U-Net \cite{ronneberger2015u} employs a symmetric encoder–decoder structure with five downsampling and upsampling stages. The encoder progressively extracts semantic features, while skip connections transfer spatial detail to the decoder, enabling precise boundary reconstruction. The input is is 1 channel (grayscale) and Output are 4 classes (background, LV endocardium, LV myocardium, LA). The feature channels are [64, 128, 256, 512, 1024]. This model serves as the baseline reference for all comparative experiments.

\subsubsection{Attention U-Net}
\begin{figure*}[!ht]
     \centering
     \includegraphics[width=1\textwidth]{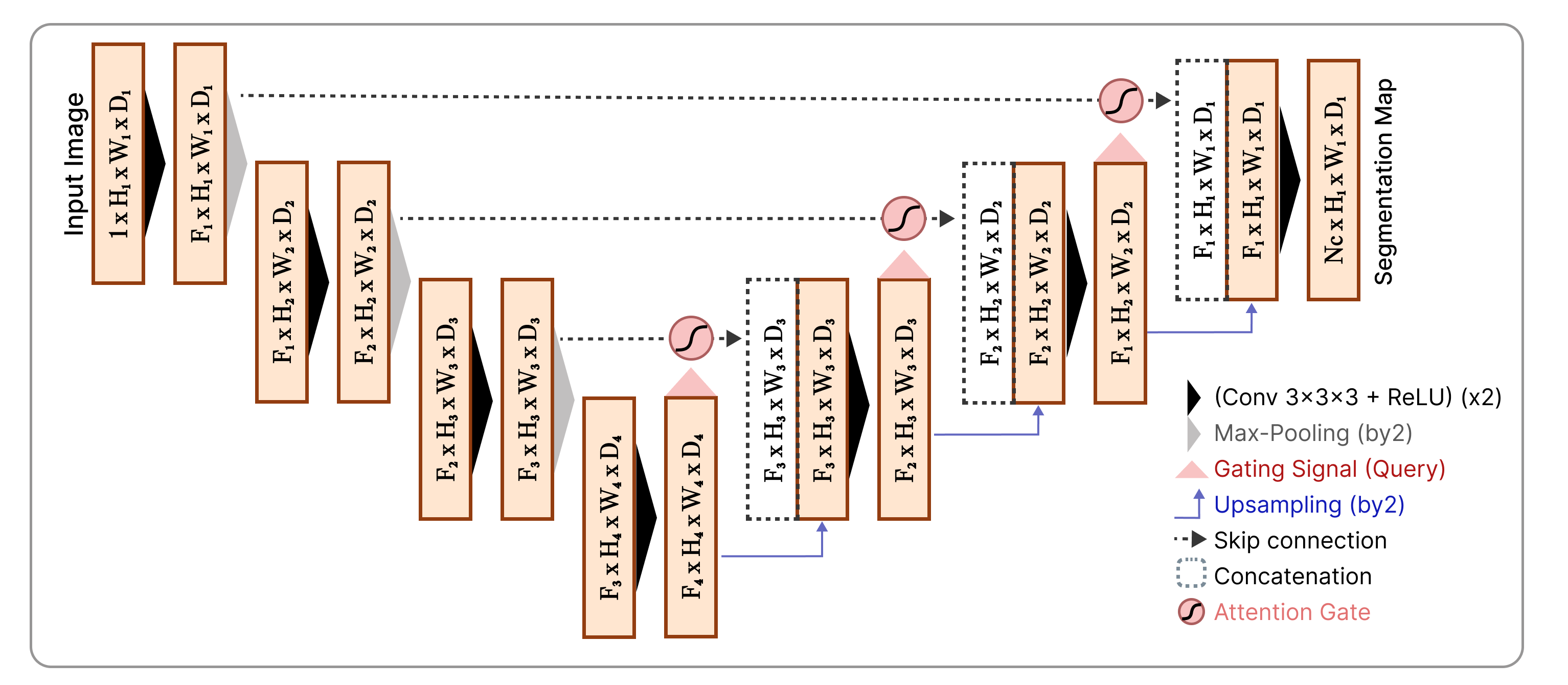}
     \caption{Architecture of the Attention U-Net model. The network extends the classical U-Net by inserting Attention Gates (red) into each skip connection. These gates receive a gating signal from the decoder and filter encoder features before concatenation, enabling the model to emphasize relevant cardiac regions (e.g., myocardium and atrial walls) and suppress background noise. Convolution, pooling, upsampling, and concatenation operations follow the standard U-Net formulation. }
     \label{attentionunet}
 \end{figure*}

The Attention U-Net \cite{oktay2018attention} enhances the original U-Net with Attention Gates (AGs) placed before skip connections as depicted in Fig. \ref{attentionunet}. These AGs selectively emphasize task-relevant regions (e.g., myocardium boundaries) and suppress irrelevant background features. In the Attention mechanism, additive gating combines encoder and decoder feature maps. Its advantage is improved segmentation of small or low-contrast regions. The AG modules are integrated at each skip level (64–512 channels).

\subsubsection{TransUNetLite}
\begin{figure*}[!ht]
     \centering
     \includegraphics[width=1\textwidth]{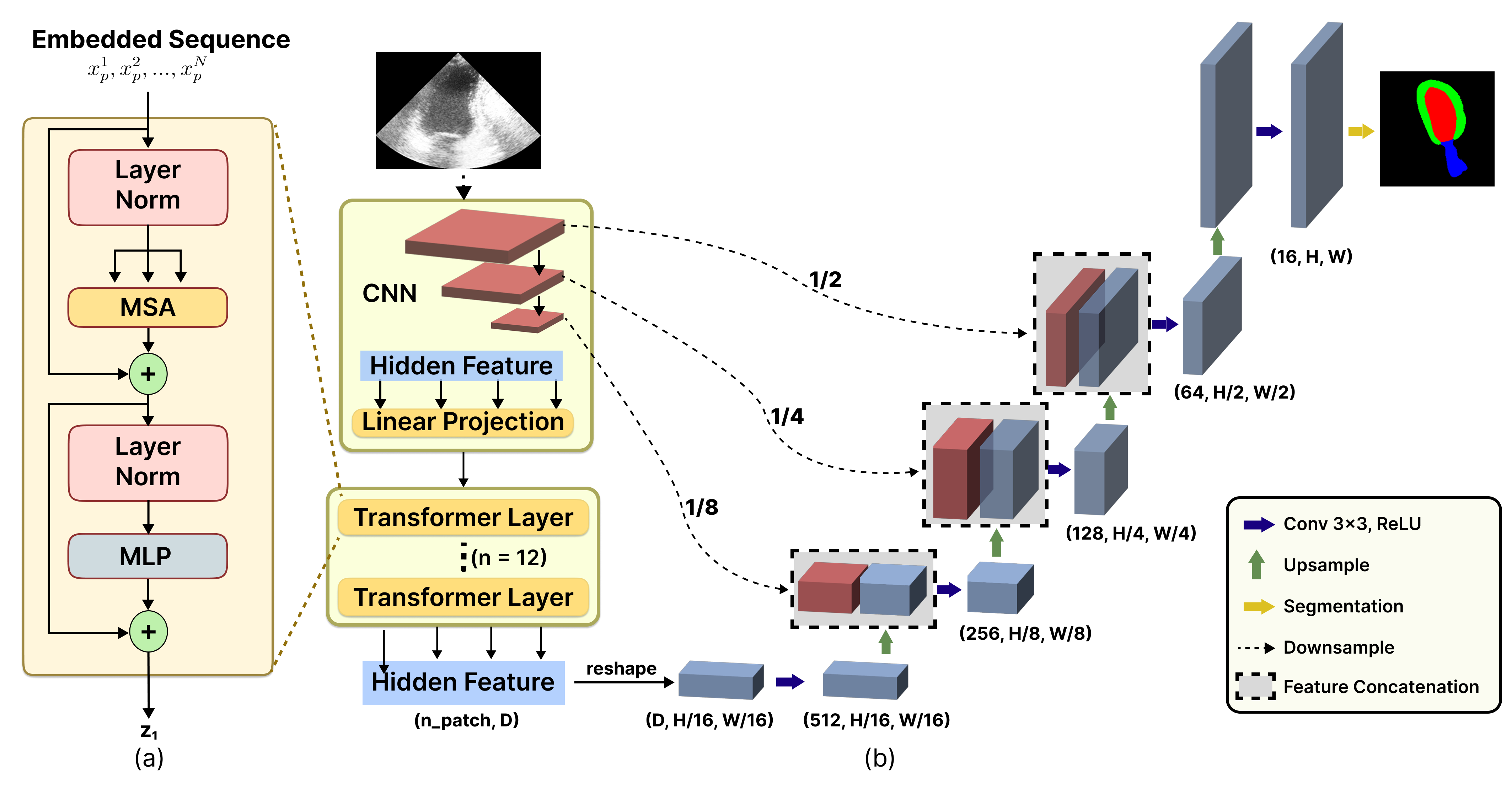}
     \caption{Overview of the TransUNet architecture \cite{chen2021transunet} for image segmentation. The model integrates a CNN encoder with a ViT to exploit both local spatial features and global contextual information. The CNN extracts hierarchical feature maps, which are linearly projected into token embeddings and processed through multiple Transformer layers. The encoded representations are reshaped and fused with multi-scale CNN features via skip connections. The decoder progressively upsamples and refines the concatenated features using convolutional blocks to generate the final segmentation map. Legends denote convolution, upsampling, downsampling, feature concatenation and segmentation operations. }
     \label{transunet}
 \end{figure*}

The TransUNetLite \cite{chen2021transunet} model combines convolutional and transformer-based components as illustrated in Fig. \ref{transunet}. The CNN encoder captures local textures, while the transformer encoder models long-range dependencies via self-attention.
The encoder in CNN performs feature extractor + patch embedding. The Transformer performs multi-head self-attention for global spatial context. The decoder in U-Net style upsampling with skip connections.The input resolution is 224$\times$224 or 512$\times$512. This hybrid approach improves contextual understanding and inter-structure coherence in cardiac segmentation.

\subsection{Training Setup}

\subsubsection{Loss Functions}
We explored three loss formulations. CE, standard pixel-wise classification loss. CE + Dice, that balances class-wise overlap and handles label imbalance. CE + Dice + Focal, which gives higher weight to hard pixels with low confidence. Empirically, CE + Dice yielded the best trade-off between stability and performance.

\subsubsection{Optimization and Regularization}
All models were trained using the Adam optimizer (learning rate = $1\times10^{-4}$, weight decay = $1\times10^{-4}$). A StepLR scheduler (decay factor 0.1 every 10 epochs) was used to adjust the learning rate. Gradient clipping ($\text{max\_norm}=1.0$) was applied to prevent gradient explosion. Training batches of 4–8 samples were used, depending on GPU capacity. Data augmentation included random flips, small rotations ($\pm10^\circ$), and intensity normalization.

\subsubsection{Resolution and Input Scaling}
Experiments were conducted at two resolutions: 256$\times$256 and 512$\times$512. While 256$\times$256 improved computational efficiency, 512$\times$512 preserved finer anatomical details and improved segmentation accuracy, especially around myocardium boundaries and atrial walls.

\subsubsection{Self-Supervised Learning (SSL)}
To utilize unlabeled data, a SimCLR-based self-supervised pretraining was performed on approximately 43,000 unlabeled PNG frames. The network learned to maximize agreement between augmented views of the same image using a contrastive objective. The pretrained encoder weights were then transferred into the U-Net and Attention U-Net backbones for fine-tuning, improving generalization and convergence stability.

\subsection{Evaluation Metrics}

\subsubsection{Quantitative Metrics}
Segmentation performance was evaluated using standard overlap-based metrics:
\begin{itemize}
    \item Dice Similarity Coefficient (Dice): measures region overlap between prediction and ground truth.
    \[
    Dice = \frac{2TP}{2TP + FP + FN}
    \]
    Computed per class: background, LV endocardium, LV myocardium, and LA.
    \item Mean Dice (mDice): arithmetic mean of Dice across all classes.
    \item Intersection over Union (IoU): evaluates the ratio of correctly predicted area to the union of prediction and ground truth:
    \[
    IoU = \frac{TP}{TP + FP + FN}
    \]
    \item Confusion Matrix: summarizes inter-class prediction errors for each epoch.
\end{itemize}

\subsubsection{Qualitative Evaluation}
Qualitative comparisons were made by overlaying predicted segmentation masks on input echocardiographic frames. These overlays illustrate model behavior visually. As shown in Fig. \ref{fig:qualitative_results}, the U-Net produces accurate global structures but sometimes blurs fine edges. Attention U-Net highlights boundaries more sharply using spatial attention. The TransUNet achieves improved inter-region consistency due to global context modeling.
These visual results complement the quantitative findings and provide intuitive insights into model segmentation quality.
In summary, this experimental framework provides a rigorous and reproducible setup for comparing three leading DL architectures for cardiac ultrasound segmentation, emphasizing the roles of data format, preprocessing, architectural design, and self-supervised pretraining on model performance.

\section{Results}\label{results}

This section presents the quantitative and qualitative findings from experiments conducted on the CAMUS dataset using U-Net, Attention U-Net, and TransUNet architectures under various preprocessing and training configurations. Results are grouped into three parts: quantitative performance (Dice and IoU metrics), qualitative visual comparisons, and ablation studies examining the effects of loss functions, resolution, and preprocessing strategies.

\subsection{Quantitative Results}

Table~\ref{tab:quantitative_results} summarizes the segmentation performance of all models across different preprocessing settings and loss configurations. All models were evaluated using the same data splits, training epochs, and evaluation metrics for fair comparison.


\begin{table}[H]
\centering
\caption{Quantitative comparison of segmentation performance (Dice coefficients in \%). U-Net, Attention U-Net, and TransUNet were trained using identical CE+Dice loss, Adam optimizer, and patient-level splits.}
\label{tab:quantitative_results}
\resizebox{\textwidth}{!}{
\begin{tabular}{lcccccc}
\toprule
\textbf{Model} & \textbf{Data Type} & \textbf{Background} & \textbf{LV Endocardium} & \textbf{LV Myocardium} & \textbf{LA} & \textbf{mDice} \\
\midrule
U-Net & NIfTI (baseline) & 98.5 & 94.2 & 93.6 & 90.8 & \textbf{94.3} \\
U-Net & PNG 16-bit & 98.1 & 91.5 & 90.8 & 88.3 & \textbf{91.0} \\
Attention U-Net & PNG 16-bit & 98.3 & 92.8 & 92.4 & 89.6 & \textbf{92.7} \\
TransUNet & PNG 16-bit & 98.4 & 93.2 & 92.0 & \textbf{91.2} & \textbf{93.2} \\
U-Net (SSL pretrained) & PNG 16-bit + SSL & 98.6 & 92.9 & 92.0 & 89.9 & \textbf{92.8} \\
\bottomrule
\end{tabular}
}
\end{table}

\paragraph{Performance trends.}
The baseline U-Net trained on NIfTI volumes achieved the highest mean Dice score of approximately 94\%, confirming that direct access to full dynamic range and uncompressed voxel intensity preserves optimal grayscale fidelity.  
When converted to 16-bit PNG format, the same model’s mean Dice dropped to 91\%, indicating that even minor quantization or intensity normalization differences can affect segmentation accuracy, especially for fine myocardial borders.

The Attention U-Net consistently outperformed the baseline U-Net by approximately 1–2\% in the myocardium and endocardium regions. This improvement is attributed to its ability to focus on salient features via attention gating, effectively suppressing background noise in echocardiographic images.

The TransUNet achieved superior performance in the LA segmentation (LA Dice $\approx$ 91.2\%), suggesting better modeling of long-range spatial dependencies due to its transformer encoder. It also demonstrated improved generalization on unseen validation cases.

Finally, incorporating self-supervised pretraining (SSL) into the U-Net encoder yielded a consistent gain of +1–2\% Dice and faster convergence during fine-tuning. The pretrained encoder provided a more stable initialization, particularly beneficial for underrepresented cardiac regions such as the apex and atrial walls.

\paragraph{Impact of preprocessing.}
Preprocessing quality strongly influenced performance. Models trained on 16-bit PNGs performed comparably to NIfTI baselines only when strict pairing, normalization, and manifest-based validation were used. Loose or inconsistent pairing led to misalignment and unstable Dice results.

\subsection{Qualitative Results}

\begin{figure}[H]
\centering
\includegraphics[width=0.95\linewidth]{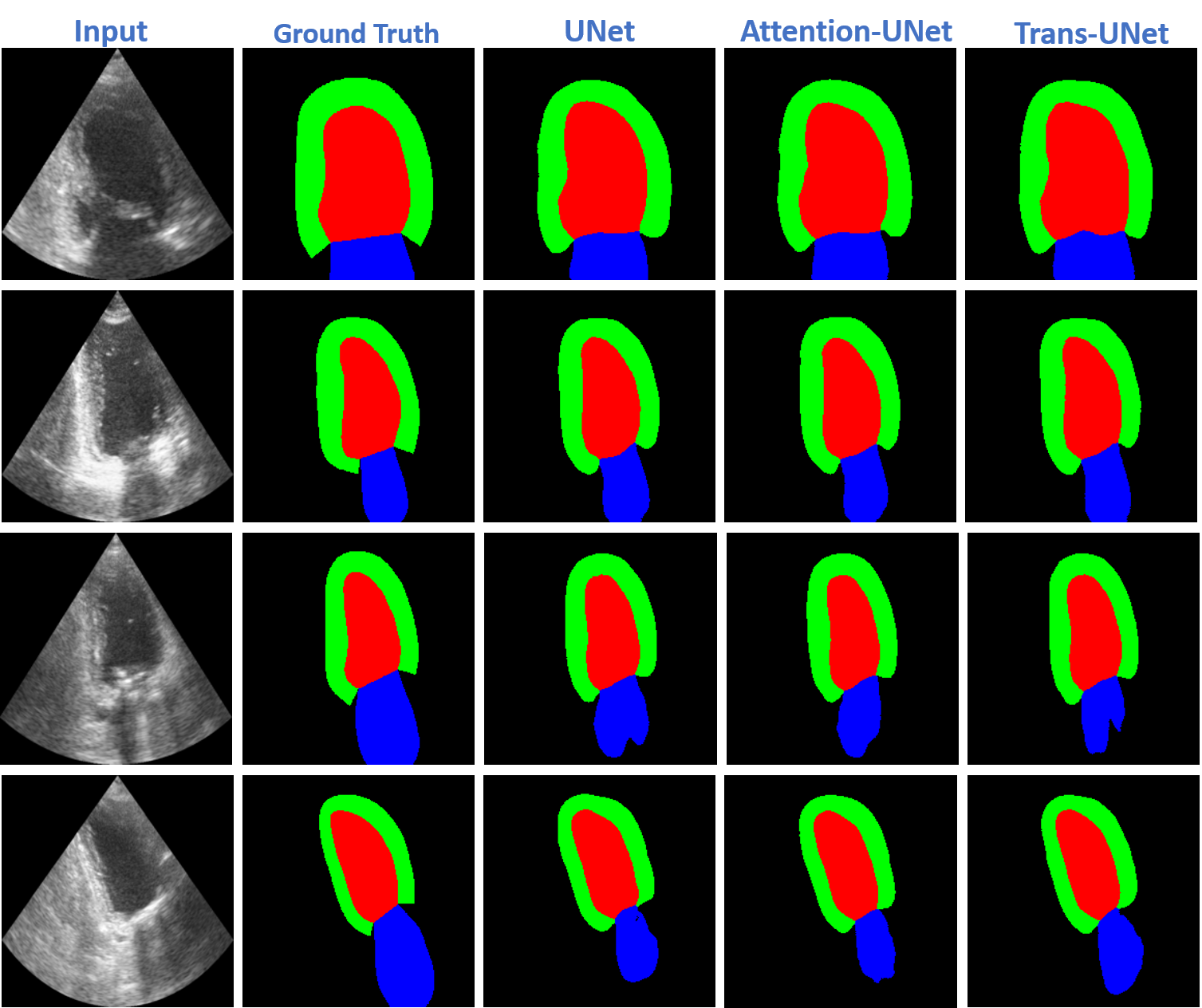}
\caption{Qualitative comparison of segmentation results on echocardiography images. The first column shows input ultrasound images, followed by ground truth segmentations. The next columns present segmentation outputs from UNet, Attention-UNet, and Trans-UNet architectures, respectively. Different anatomical regions are indicated by distinct colors, allowing for visual assessment of segmentation accuracy across models.}
\label{fig:qualitative_results}
\end{figure}

Qualitative comparisons were made by overlaying predicted segmentation masks on the original echocardiographic frames. Representative examples are shown in Figure~\ref{fig:qualitative_results}.

\paragraph{Visual interpretation.}
Visual inspection confirms that the U-Net delineates the general cardiac structures accurately but occasionally exhibits blurred myocardium boundaries and small leakage near the apex, especially in low-contrast frames.

The Attention U-Net produces sharper boundaries with improved localization around the endocardium and myocardium interfaces. The attention maps highlight the ventricular and atrial walls more precisely, reducing false positives in noisy regions.

The TransUNet demonstrates superior shape regularity and smoother contours for the LA. Its transformer encoder enables better global context awareness, preventing discontinuities and ensuring consistent segmentation across spatially distant regions.

Overall, the qualitative results align with the quantitative analysis, confirming that attention and transformer-based mechanisms enhance structural integrity and region focus in challenging echocardiographic frames.

\subsection{Ablation Studies}

To investigate the effects of architectural and training variations, ablation experiments were conducted across three dimensions such as loss function combinations, input resolution, and preprocessing rigor. The corresponding mean Dice results are summarized in Table~\ref{tab:ablation_results}.

\begin{table}[H]
\centering
\caption{Ablation study on loss functions, image resolution, and preprocessing strategies (mean Dice \%).}
\label{tab:ablation_results}
\begin{tabular}{lccc}
\toprule
\textbf{Configuration} & \textbf{U-Net} & \textbf{Attention U-Net} & \textbf{TransUNet} \\
\midrule
CE loss only & 89.4 & 90.1 & 91.0 \\
CE + Dice loss & \textbf{91.0} & \textbf{92.7} & \textbf{93.2} \\
CE + Dice + Focal loss & 90.8 & 92.1 & 92.9 \\
\midrule
Resolution 256$\times$256 & 90.5 & 91.8 & 92.4 \\
Resolution 512$\times$512 & \textbf{91.0} & \textbf{92.7} & \textbf{93.2} \\
\midrule
Loose PNG preprocessing & 88.7 & 89.9 & 91.1 \\
Strict manifest-based preprocessing & \textbf{91.0} & \textbf{92.7} & \textbf{93.2} \\
\bottomrule
\end{tabular}
\end{table}

\paragraph{Loss functions.}
The combination of CE and Dice losses provided the best trade-off between convergence stability and region overlap accuracy. Adding Focal loss slightly improved hard-pixel learning but introduced mild training instability for small batch sizes.

\paragraph{Image resolution.}
Increasing the input resolution from 256$\times$256 to 512$\times$512 led to a 0.5–1\% improvement in Dice, primarily for myocardium segmentation, due to enhanced boundary preservation.

\paragraph{Preprocessing strictness.}
Strict manifest-based preprocessing (exact image–mask pairing and normalization) produced consistent Dice improvements of 2–3\% compared to loosely paired PNG datasets, confirming the necessity of rigorous data curation for reproducibility.

Overall, the experiments demonstrate that U-Net remains a strong and reliable baseline, although its performance is notably affected by variations in intensity normalization. Attention U-Net enhances fine structural delineation by incorporating spatial gating mechanisms, leading to more precise boundary detection. TransUNet achieves the highest overall generalization and shape consistency, benefiting from its hybrid CNN-transformer architecture. Furthermore, the quality of preprocessing and the use of self-supervised pretraining play a crucial role in determining segmentation accuracy and model robustness. These results establish a reproducible benchmark for cardiac ultrasound segmentation, highlighting how preprocessing, architectural design, and self-supervision jointly determine the accuracy and stability of DL models on the CAMUS dataset.

\section{Discussion}\label{discussion}

The conducted experiments provide a comprehensive understanding of how different data preprocessing strategies and architectural choices influence cardiac segmentation performance on the CAMUS dataset. In this section, we discuss key findings, practical insights, and potential directions for future research.

\subsection{Key Observations}

\paragraph{Data Representation and Fidelity.}
One of the most consistent findings is that the NIfTI-based pipeline achieved the highest mean Dice score (approximately 94\%), outperforming all PNG-based configurations. This improvement is primarily attributed to the fact that NIfTI files preserve the original floating-point intensity values without quantization or interpolation, ensuring higher grayscale fidelity and more stable normalization across samples.  

In contrast, conversion to 16-bit PNG format introduced minor intensity rounding and spatial interpolation artifacts, leading to a 2–3\% drop in Dice coefficient. These effects were particularly pronounced in low-contrast myocardial regions, where subtle edge information was lost during conversion.

\paragraph{Model Architecture.}
The architectural comparison revealed complementary strengths among the three evaluated models. The U-Net remains a strong baseline, providing efficient and accurate segmentation across all cardiac structures. However, its performance can degrade on small or ambiguous regions due to its limited ability to prioritize relevant spatial features.

The Attention U-Net addressed this limitation by integrating attention gates before skip connections. These modules adaptively weighted encoder features, resulting in improved segmentation of small regions, especially near the ventricular apex and thin myocardium borders. The addition of attention yielded consistent improvements of 1–2\% Dice in localized regions.

The TransUNet model demonstrated superior generalization and structural coherence. Its hybrid design, combining convolutional feature extraction with transformer-based self-attention, captured both local details and global spatial dependencies. This enabled smoother and more anatomically consistent delineations, particularly for the LA, which exhibited the highest class-specific Dice improvements.

\paragraph{Training Behavior and Losses.}
Among the tested loss functions, the combination of CE and Dice losses consistently yielded the best trade-off between convergence stability and accuracy. This hybrid formulation balances pixel-wise classification with region-level overlap, effectively mitigating class imbalance between cardiac structures. The addition of Focal loss provided minor gains on small datasets but introduced occasional instability during backpropagation.

\subsection{Practical Lessons Learned}

From a methodological standpoint, several implementation and preprocessing practices proved essential for achieving stable and reproducible results, such as Label Integrity: after each preprocessing or format conversion, all label masks should be verified to contain only the valid class indices (0–3). Spurious values from interpolation or compression can cause unpredictable model behavior. Resolution Scaling: increasing the input resolution from 256$\times$256 to 512$\times$512 significantly improved Dice scores for the myocardium class. The higher resolution preserved finer boundary structures and improved edge continuity in segmentation masks. Loss Configuration: the CE+Dice loss combination matched or exceeded the best results reported in the literature for the CAMUS dataset, suggesting that sophisticated hybrid loss functions are not always necessary when paired with robust preprocessing and normalization. Preprocessing Strictness: strict manifest-based dataset pairing was critical for ensuring one-to-one correspondence between images and masks. Even a small number of misaligned pairs caused unstable training and reduced performance consistency. These lessons underscore the importance of rigorous dataset curation, proper normalization, and architectural selection tailored to data modality.

\subsection{Future Directions}

The current findings open several promising directions for extending this work both methodologically and clinically:

\paragraph{1. Self-Supervised and Semi-Supervised Pretraining.}
Self-supervised pretraining techniques, such as SimCLR and Masked Autoencoders (MAE), can leverage the vast amount of unlabeled echocardiography slices available in the CAMUS dataset. By learning generalizable representations from unlabeled data, models can achieve faster convergence, improved robustness, and better adaptation to small labeled datasets.

\paragraph{2. GPT-based Multimodal Labeling and Error Analysis.}
Recent multimodal GPT systems capable of image understanding offer potential for pseudo-label generation and post-training error analysis. Large language–vision models can automatically identify segmentation inconsistencies, generate polygon annotations, and provide text-based rationales for errors (e.g., misclassified myocardium edges). This hybrid human–AI collaboration could accelerate dataset expansion and quality control.

\paragraph{3. Clinical Integration and Workflow Automation.}
Beyond segmentation performance, future work should focus on clinical applicability, such as integrating segmentation outputs into automated diagnostic pipelines for estimating cardiac function metrics (e.g., EF, ED/systolic volumes). Deploying these models into real-time echocardiographic workflows would facilitate decision support in cardiology practice, enabling faster and more reproducible cardiac assessments.

Overall, this study demonstrates that the fidelity of input data (NIfTI vs. PNG) substantially impacts segmentation accuracy. Whereas attention mechanisms improve small-region sensitivity, while transformer-based encoders enhance global structural consistency. Likewise, proper preprocessing, high-resolution inputs, and hybrid CE+Dice loss are essential for robust model performance. Similarly, incorporating self-supervised and GPT-based multimodal approaches represents a natural next step toward scalable, clinically integrated cardiac image segmentation. These findings collectively contribute to establishing a reproducible benchmark for comparing DL architectures in cardiac ultrasound segmentation and offer practical guidance for future research in medical image analysis.

\section{Conclusion and Future Directions}\label{con}

This paper presented a combined review and experimental benchmark for cardiac ultrasound segmentation, covering existing literature, public datasets, and three widely used architectures. While earlier studies summarized DL methods for cardiovascular imaging, their results are difficult to compare due to differences in datasets, preprocessing, and evaluation protocols. This work addresses that gap by reviewing the field and providing a controlled, reproducible benchmark. Using the CAMUS dataset, we evaluated U-Net, Attention U-Net, and TransUNet under matched preprocessing and training settings. U-Net remained a strong baseline, Attention U-Net improved boundary detail in challenging regions, and TransUNet showed the best global shape consistency thanks to its CNN–Transformer hybrid design. Preprocessing quality proved critical: training directly on NIfTI data achieved Dice scores up to 94\%, while 16-bit PNG conversion produced a small performance drop. Beyond supervised learning, two strategies helped make better use of unlabeled data. Self-supervised SimCLR pretraining improved encoder stability, and SAM-based pseudo-labeling expanded the training set with confidence-filtered masks. These results show that unlabeled frames can meaningfully strengthen segmentation performance. Overall, this work provides (i) a concise review of cardiac segmentation research, (ii) a fair benchmark of three representative architectures, and (iii) practical evidence that preprocessing rigor and intelligent data utilization are as important as model choice. Future directions include hybrid CNN–Transformer designs, multimodal GPT-based annotation tools, and semi-supervised workflows that combine human input with foundation models.

\section*{CRediT authorship contribution statement}
\textbf{Zahid Ullah:} Conceptualization, Methodology, Software, Formal analysis, Investigation, Data curation, Writing - original draft, Writing - review \& editing. \textbf{Muhammad Hilal:} writing Original draft. \textbf{Eunsoo Lee:} Writing original draft. \textbf{Dragan Pamucar:} Formal analysis, investigation, Supervision. \textbf{Jihie Kim:} Formal analysis, Investigation, Supervision, Project administration.

\section*{\textbf{Declaration of Competing Interests}} The authors declare that they have no known competing financial interests or personal relationships that could have appeared to influence the work reported in this paper.

\section*{Acknowledgements}
This research was supported by the MSIT(Ministry of Science and ICT), Korea, under the ITRC(Information Technology Research Center) support program(IITP-2026-RS-2020-II201789), and the Artificial Intelligence Convergence Innovation Human Resources Development(IITP-2026-RS-2023-00254592) supervised by the IITP(Institute for Information \& Communications Technology Planning \& Evaluation).

\bibliography{sample}

\end{document}